\PassOptionsToPackage{dvipsnames,table}{xcolor}

\documentclass[acmsmall,screen]{acmart}
\AtBeginDocument{%
  }

\usepackage{amsmath,amsfonts,bm}
\usepackage[ruled,linesnumbered]{algorithm2e}
\usepackage{array}
\usepackage{textcomp}
\usepackage{stfloats}
\usepackage{url}
\usepackage{verbatim}
\usepackage{graphicx}

\usepackage{multirow,booktabs,color,soul,threeparttable}
\usepackage{rotating}
\usepackage{textcomp,tgpagella} 

\usepackage{lscape}
\usepackage{listings}
\usepackage{colortbl}
\usepackage{epstopdf}
\usepackage{multirow}
\usepackage{float}
\usepackage[T1]{fontenc}
\usepackage{subfigure}
\usepackage{times}

\usepackage{xr}
\usepackage{xr-hyper}
\usepackage{hyperref}

\newcommand{\ie}{\mbox{\emph{i.e.}}}
\newcommand{\eg}{\mbox{\emph{e.g.}}}

\definecolor{hl}{rgb}{0.75,0.75,0.75}
\definecolor{hl2}{rgb}{0.92,0.92,0.92}
\sethlcolor{hl}

\usepackage{DejaVuSansMono}

\newcommand{\code}[1]{%
  {\colorbox{gray!10}{\scriptsize\texttt{#1}}}%
}

\graphicspath{{/}{figs/}}

\begin{document}

\title{MToP: A MATLAB {Benchmarking} Platform for Evolutionary Multitasking}

\author{Yanchi~Li}
\email{int\_lyc@cug.edu.cn}
\author{Wenyin~Gong}
\authornote{Corresponding author.}
\email{wygong@cug.edu.cn}
\author{Tingyu~Zhang}
\affiliation{%
  \institution{School of Computer Science, China University of Geosciences}
  \city{Wuhan}
  \state{Hubei}
  \country{China}
}

\author{Fei~Ming}
\affiliation{%
  \institution{Trustworthy and General Artificial Intelligence Laboratory, School of Engineering, Westlake University}
  \city{Hangzhou}
  \state{Zhejiang}
  \country{China}
}

\author{Shuijia~Li}
\affiliation{%
  \institution{College of Systems Engineering, National University of Defense Technology}
  \city{Changsha}
  \state{Hunan}
  \country{China}
}

\author{Qiong~Gu}
\affiliation{%
  \institution{School of Computer Engineering, Hubei University of Arts and Science}
  \city{Xiangyang}
  \state{Hubei}
  \country{China}
}

\author{Yew-Soon~Ong}
\affiliation{%
  \institution{College of Computing and Data Science, Nanyang Technological University}
  \city{Singapore}
  \country{Singapore}
}
\affiliation{%
  \institution{Centre for Frontier AI Research (CFAR), Agency for Science, Technology and Research}
  \city{Singapore}
  \country{Singapore}
}

\renewcommand{\shortauthors}{Li et al.}

\begin{abstract}
  Evolutionary multitasking (EMT) has emerged as a popular topic of evolutionary computation over the past decade. It aims to concurrently address multiple optimization tasks within limited computing resources, leveraging inter-task knowledge transfer techniques. Despite the abundance of multitask evolutionary algorithms (MTEAs) proposed for multitask optimization (MTO), there remains a need for a comprehensive software platform to help researchers evaluate MTEA performance on benchmark MTO problems as well as explore real-world applications. To bridge this gap, we introduce the first open-source benchmarking platform, named MToP, for EMT. MToP incorporates over 50 MTEAs, more than 200 MTO problem cases with real-world applications, and over 20 performance metrics. Based on these, we provide benchmarking recommendations tailored for different MTO scenarios. Moreover, to facilitate comparative analyses between MTEAs and traditional evolutionary algorithms, we adapted over 50 popular single-task evolutionary algorithms to address MTO problems. Notably, we release extensive pre-run experimental data on benchmark suites to enhance reproducibility and reduce computational overhead for researchers. MToP features a user-friendly graphical interface, facilitating results analysis, data export, and schematic visualization. More importantly, MToP is designed with extensibility in mind, allowing users to develop new algorithms and tackle emerging problem domains. The source code of MToP is available at: \href{https://github.com/intLyc/MTO-Platform}{\color{violet}\footnotesize\texttt{https://github.com/intLyc/MTO-Platform}}
\end{abstract}





\maketitle

\section{Introduction}
Evolutionary computation (EC), inspired by natural evolution, has experienced rapid growth owing to its effectiveness and efficiency. Evolutionary algorithms (EAs), the cornerstone of EC, have demonstrated remarkable success in addressing black-box optimization problems due to their robustness and ease of implementation. Researchers have dedicated significant efforts to designing tailored EAs for various complex black-box optimization problems, including constrained optimization~\cite{Wang2020CORCO, Tian2021CCMO}, multi-objective optimization~\cite{Zhang2007MOEA-D, Deb2014NSGA3}, and combinatorial optimization~\cite{Feng2021VRPHTO, Feng2021EEMTA}. In recent years, driven by escalating computational demands and the emergence of cloud computing, there has been a growing emphasis on utilizing EAs to tackle multiple optimization tasks concurrently, known as evolutionary multitasking~\cite{Wei2021Review-MTO}. A multitask optimization (MTO) problem within the realm of EMT, comprising $K$ {minimization} tasks, aims to find solutions $(\bm{x}_{1}^\ast,\bm{x}_{2}^\ast,...,\bm{x}_{K}^\ast)$ for all tasks, which can be formulated as follows:
\begin{equation}\label{eq:def-mtop}
  \begin{gathered}
    (\bm{x}_{1}^\ast,\bm{x}_{2}^\ast,...,\bm{x}_{K}^\ast)  = \arg\min \Big[F_1(\bm{x}_1),F_2(\bm{x}_2),...,\bm{F}_K(\bm{x}_K)\Big], \\
    \textit{s.t. } \bm{x}_k \in \Omega_k,\ k=1,2,...,K,\\
  \end{gathered}
\end{equation}
where $\Omega_k$ and $\bm{F}_k$ are the decision space and objective function of the $k$-th task. Note that for a {multi-objective multi-task optimization problem}, the $\bm{F}_k$ contains multiple objective functions $(f_1,f_2,...,f_{M_k})$ and $\bm{x}_{k}^\ast$ becomes a set of Pareto sets for the $k$-th task. {For task $k$ with upper and lower bounds $\bm{L}_k$ and $\bm{U}_k$ and dimension $D_k$, the solution $\bm{x}$ of its decision space is typically mapped to the unified search space~\cite{Gupta2016MFEA} as follows:
    \begin{equation}\label{eq:uni-space}
      \bm{x}^\prime = \frac{\bm{x} - \bm{L}_k}{\bm{U}_k-\bm{L}_k}.
    \end{equation}
    In addition, the dimensionality of task $k$ is expanded to $\max(D_1,...,D_K)$. During {objective} function evaluation, dimensions and upper and lower bounds are linearly reproduced without loss of precision.} EMT has found successful applications in various domains, including engineering scheduling~\cite{Feng2021VRPHTO, Gupta2021Half}, nonlinear equation systems~\cite{Li2024Evolutionary}, feature selection in machine learning~\cite{Chen2021Evolutionary}, anomaly detection~\cite{Wang2022AUC}, point cloud registration~\cite{Wu2024Evolutionary, Wu2023Evolutionary}, and reinforcement learning~\cite{Zhang2023Multitask}.

To expedite and enhance the concurrent resolution of multiple optimization tasks, researchers have endeavored to leverage task similarity to augment EAs with knowledge extraction and transfer techniques~\cite{Tan2023Knowledge, Gupta2018Insights}. Through knowledge transfer, EAs can effectively exploit implicit parallelism to achieve superior solutions across multiple tasks while conserving computational resources~\cite{Tan2021ETO, Li2024MFEA-RL}. The first attempt of EMT can be traced back to the multifactorial EA~\cite{Gupta2016MFEA}, which introduced an implicit knowledge representation via random mating among optimization tasks. Subsequently, numerous EAs tailored for MTO have emerged. These multitask evolutionary algorithms (MTEAs) adopt either a multifactorial framework utilizing a single population for multiple tasks~\cite{Bali2020MFEA2, Zhou2021MFEA-AKT}, or a multi-population framework allocating separate populations for each task~\cite{Li2020MFMP, Li2023MTEA-SaO, Li2024MTES-KG}. Moreover, to facilitate decision space mapping across different tasks, various techniques such as unified search space~\cite{Gupta2016MFEA, Zhou2021MFEA-AKT}, auto-encoding~\cite{Feng2019EMEA, Zhou2021Learnable, Gu2025MTEA-PAE}, affine transformation~\cite{Xue2022AT-MFEA, Lin2024Ensemble}, and adversarial generative models~\cite{Liang2023EMT-GS} have been proposed in MTEAs. Given the significant impact of knowledge transfer on solving MTO problems, MTEAs with adaptive control strategies for knowledge transfer, such as similarity judgment~\cite{Bali2020MFEA2, Jiang2023BoKT}, knowledge selection~\cite{Wang2022MTEA-AD, Liang2022EMaTO-MKT, Li2025MTDE-MKTA}, and historical feedback~\cite{Li2022CompetitiveMTO, Lin2021EMT-ET, Li2023MTSRA}, have also been investigated. Despite the proliferation of MTEAs proposed by researchers, there is currently no standardized programming language, code pattern, or software platform for EMT source codes. This presents challenges for newcomers entering the field of EMT and for researchers seeking to conduct convenient experimental comparisons of algorithms.

To overcome these challenges, we present MToP, an open-source MATLAB platform tailored for advancing the EMT field. MToP is designed to provide a comprehensive software platform for researchers to evaluate MTEAs on benchmark MTO problems and explore real-world applications. {The selection of MATLAB is based on several strategic factors. First, many foundational and seminal works in the EMT field, including the original MFEA~\cite{Gupta2016MFEA} and MFEA-II~\cite{Bali2020MFEA2}, were developed and open-sourced in MATLAB. This established a strong precedent and a rich ecosystem of existing code for the community. Second, the prominence of other successful EC platforms in MATLAB, most notably PlatEMO~\cite{Tian2017PlatEMO}, provides a high-quality open-source reference. This synergy allows for the adaptaion and reuse of well-vetted modules, such as those for multi-objective optimization, which is crucial for the multi-objective MTO subfield. Finally, MATLAB's powerful numerical computing environment and its capabilities for building user-friendly graphical interfaces make it an ideal choice for a comprehensive platform.} The main contributions of MToP are summarized as follows:
\begin{enumerate}
  \item MToP features a user-friendly graphical user interface (GUI) comprising test, experiment, and data-process modules. These modules facilitate researchers in understanding problem characteristics, conducting comparative experiments, solving problems in parallel, statistically analyzing results, plotting result figures, and managing experimental data. Moreover, MToP offers a modular implementation of algorithms, problems and performance metrics. On top of these, it provides an extensive public application programming interface (API) with template functions for population initialization, function evaluation, evolutionary operators, and environmental selection.
  \item MToP encompasses a wide array of algorithms, problems, and metrics, all accessible via a public API. Over 50 MTEAs are implemented, catering to single-objective, multi-objective, constrained, many-task, and competitive multitask types. Additionally, to facilitate comparative analyses between MTEAs and popular traditional EAs, MToP integrates more than 50 single-task EAs of diverse types. In terms of synthetic test problems, MToP incorporates over 200 benchmark MTO problems alongside several real-world applications. Lastly, MToP features a variety of multitask and single-task performance metrics, such as multitask score, objective value, hypervolume, and running time, providing comprehensive evaluation capabilities. {We also release extensive pre-run experimental data to enhance reproducibility and reduce computational overhead for researchers.}
  \item MToP is designed to be easily extended, allowing for the seamless addition of new algorithms, problems, and metrics. By adhering to established coding patterns and implementing functionality based on the public API, new code can be seamlessly integrated and utilized within MToP. Given its status as a completely open-source project, researchers have the opportunity to leverage existing algorithms, problems, and metrics as the foundation for novel ideas. Through the collaborative platform GitHub, MToP undergoes continuous updates and enhancements, ensuring that it remains at the forefront of EMT research.
\end{enumerate}

The rest of this paper is organized as follows. {Section~\ref{sec:related} reviews related software platforms in EC and justifies the need for MToP.} Section~\ref{sec:arch} elaborates on the architecture of MToP. {Section~\ref{sec:guide} provides guidelines for using MToP. Section~\ref{sec:validate} presents the experimental validation of MToP. Finally, Section~\ref{sec:discuss} gives discussions and outlook of MToP.}

\section{Related Work}\label{sec:related}

{
  As the field of EMT continues to gain momentum, there is an urgent need for a convenient and user-friendly software platform to facilitate the benchmarking of MTEAs. Furthermore, accessible source code and platforms are indispensable for exploring the real-world applications of the EMT field. Open-source and user-friendly software platforms play a crucial role in fostering the advancement of a research field. In the field of EC, several popular and successful software platforms have significantly contributed to the development of EAs and evolutionary optimization:
  \begin{itemize}
    \item \textbf{PlatEMO}~\cite{Tian2017PlatEMO}: A comprehensive MATLAB platform tailored for evolutionary multi-objective optimization.
    \item \textbf{EDOLAB}~\cite{Peng2023EDOLAB}: A MATLAB-based platform focusing on evolutionary dynamic optimization.
    \item \textbf{IOHprofiler}~\cite{Doerr2018IOHprofiler}: A modular framework, including IOHexperimenter~\cite{deNobel2024IOHexperimenter} and IOHanalyzer~\cite{Wang2022IOHanalyzer}, for the detailed benchmarking and analysis of iterative optimization heuristics, with a primary focus on single-objective and multi-objective problems.
    \item \textbf{EvoX}~\cite{Huang2024EvoX}: A distributed GPU-accelerated library focused on scalability and expediting complex optimization and reinforcement learning tasks.
    \item \textbf{PyPop7}~\cite{Duan2024PyPop7}: A pure-Python library for population-based single-objective black-box optimization, particularly for large-scale optimization.
    \item \textbf{MetaBox}~\cite{Ma2023MetaBox}: A Python-based platform for meta-black-box optimization, which uses meta-learning to design optimizers, with its latest version using MTO as a training scenario.
  \end{itemize}
}

\begin{table}[htbp]
  \centering
  \renewcommand{\arraystretch}{1.2}
  \caption{Comparison of core focus areas among popular evolutionary computation platforms.}
  \resizebox{\textwidth}{!}{
    \begin{tabular}{lccc}
      \toprule
      Platform                                & Language     & Core Focus                                     & MTO Support             \\
      \midrule
      PlatEMO~\cite{Tian2017PlatEMO}          & MATLAB       & Evolutionary Multi-objective Optimization      & Partial (v4)            \\
      EDOLAB~\cite{Peng2023EDOLAB}            & MATLAB       & Evolutionary Dynamic Optimization              & No                      \\
      IOHprofiler~\cite{Doerr2018IOHprofiler} & C++/Python/R & Benchmarking Iterative Optimization Heuristics & No                      \\
      EvoX~\cite{Huang2024EvoX}               & Python       & Distributed GPU-accelerated Optimization       & No                      \\
      PyPop7~\cite{Duan2024PyPop7}            & Python       & Large-scale Black-box Optimization             & No                      \\
      MetaBox~\cite{Ma2023MetaBox}            & Python       & Meta-black-box Optimization via Meta-learning  & As a test scenario (v2) \\
      MToP (This work)                        & MATLAB       & Evolutionary Multitask Optimization            & Yes (Native)            \\
      \bottomrule
    \end{tabular}}
  \label{tab:platforms}
\end{table}

{
A review of these tools, summarized in Table~\ref{tab:platforms}, justifies the development of a new, dedicated framework. At the time MToP development commenced\footnote{MToP has been continuously updated since Sep. 21, 2021 at: \href{https://github.com/intLyc/MTO-Platform}{\color{violet}\scriptsize\texttt{https://github.com/intLyc/MTO-Platform}}}, a platform with native, comprehensive support for MTO benchmarking was absent. The most prominent existing platform, PlatEMO~\cite{Tian2017PlatEMO}, has a core architecture that is multi-objective optimization-centric, meticulously designed for solving single-task multi-objective optimization. This presents a fundamental architectural mismatch for EMT. Unlike the traditional single-task EC field, solving MTO problems with MTEAs necessitates the simultaneous evolution of multiple optimization tasks, the implementation of inter-task solution space mapping, and diverse knowledge transfer techniques. Additionally, the performance metrics in EMT are diverse, encompassing both single-task and multitask metrics, which brings uncertainty to the pattern of results display and analysis. These distinctive requirements present significant challenges for implementation in platforms not designed for them. Consequently, extending current framework like PlatEMO to natively handle these MTO-specific concepts is non-trivial and leads to significant extensibility limitations. This is particularly evident in the context of many-task optimization, where PlatEMO's support for specialized multitask metrics is restricted. Although preliminary MTO support was added to PlatEMO (v4.0)\footnote{PlatEMO v4.0 (released on Oct. 13, 2022) introduced preliminary MTO support.}, these additions are supplementary rather than architecturally native. The other platforms listed in Table~\ref{tab:platforms} are similarly specialized for different, non-MTO-centric goals. Moreover, EMT has been extended to the subfields of many-task optimization~\cite{Chen2020MaTDE, Liaw2019SBO, Li2024TNG-NES}, multi-objective MTO~\cite{Gupta2017MO-MFEA, Li2025MTEA-DCK}, competitive MTO~\cite{Li2022CompetitiveMTO, Li2023MTSRA, Li2024CMO-MTO}, and constrained multi-task optimization~\cite{Li2022CMTO-Benchmark, Zhang2024CEDA}. This growing complexity further necessitates a dedicated, MTO-native framework. While a new framework was deemed necessary, MToP actively integrates and appropriately cites code modules from these established platforms where feasible, thereby reducing redundant development and focusing on its core MTO-specific contributions.
}

\section{Architecture of MToP}\label{sec:arch}

{
  In this section, we present the architecture of MToP, encompassing its functional modules, project structure, and code patterns.
}

\subsection{Functional Modules}\label{sec:arch:func}

MToP is organized into three interconnected functional modules to support a full research workflow. Researchers can first use the \code{Test Module} for preliminary analysis and visualization of specific algorithms and problems. The \code{Experiment Module} facilitates large-scale, multi-run comparative experiments. Finally, the \code{Data Process Module} provides tools to manage the datasets generated by the \code{Experiment Module}, such as merging or splitting results for flexible post-processing. The following subsections detail the architectural role and capabilities of these modules. A practical guide on their specific operation and usage is provided in Section~\ref{sec:guide}.

{
Notably, MToP is architecturally designed without any inherent software limitation on the number of tasks. Users can programmatically define benchmarks with an arbitrary number of tasks without hard-coded limits.
}

\subsubsection{Test Module}\label{sec:arch:func:test}

\begin{figure*}[htbp]
  \centering
  \subfigure[Function landscapes (1D)]{\includegraphics[scale=0.3]{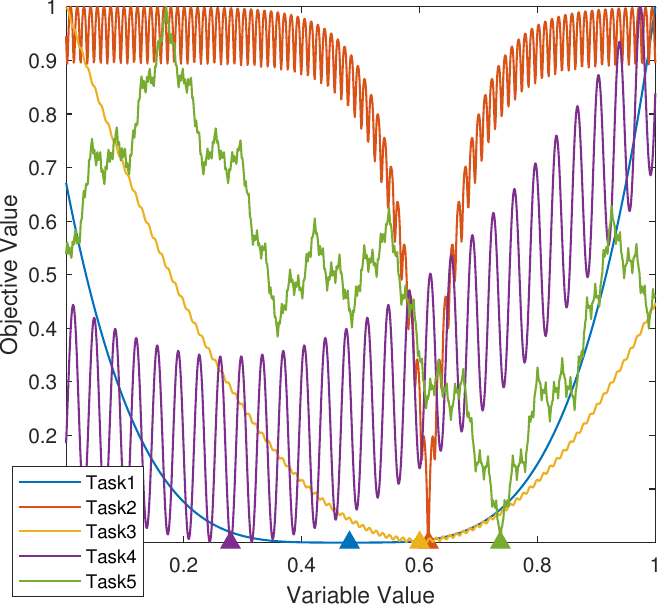}}
  \subfigure[Function landscapes (2D)]{\includegraphics[scale=0.22]{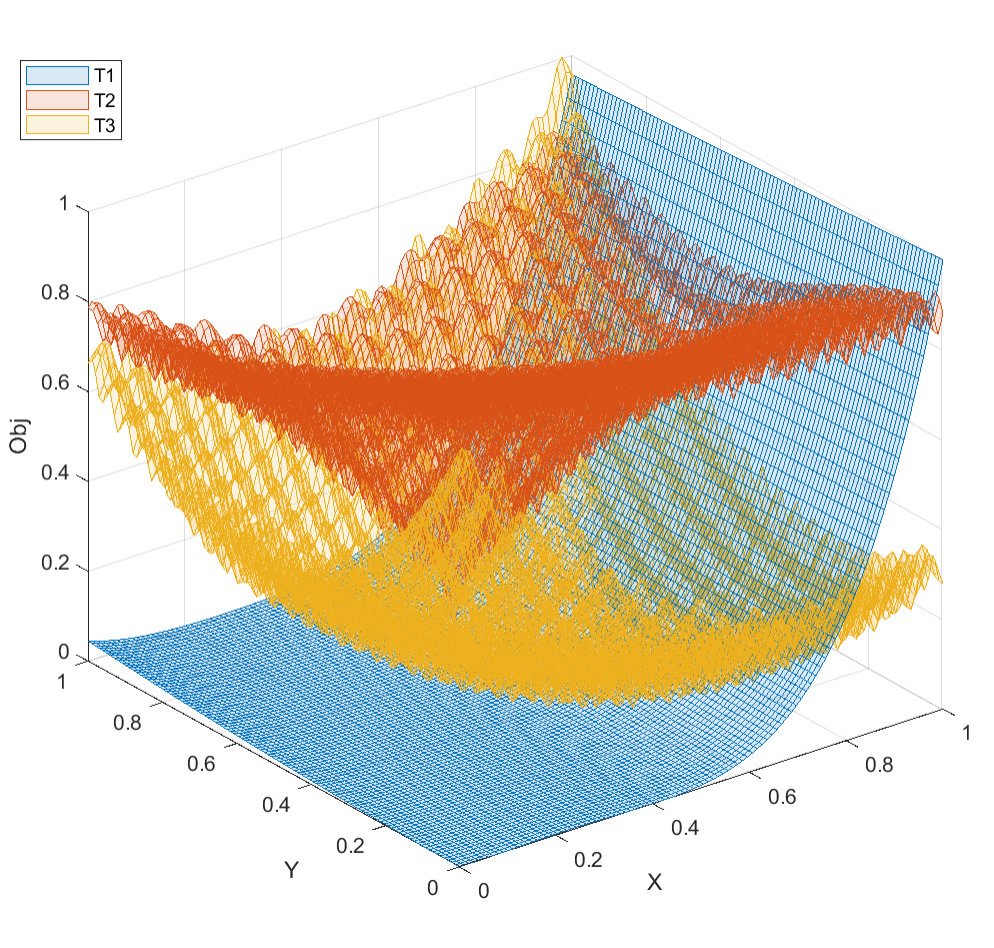}}
  \subfigure[Feasible regions]{\includegraphics[scale=0.43]{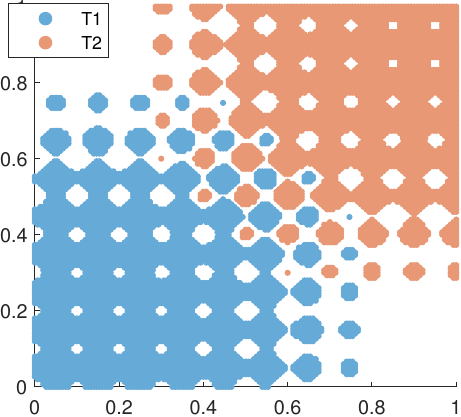}}
  \subfigure[Pareto front]{\includegraphics[scale=0.095]{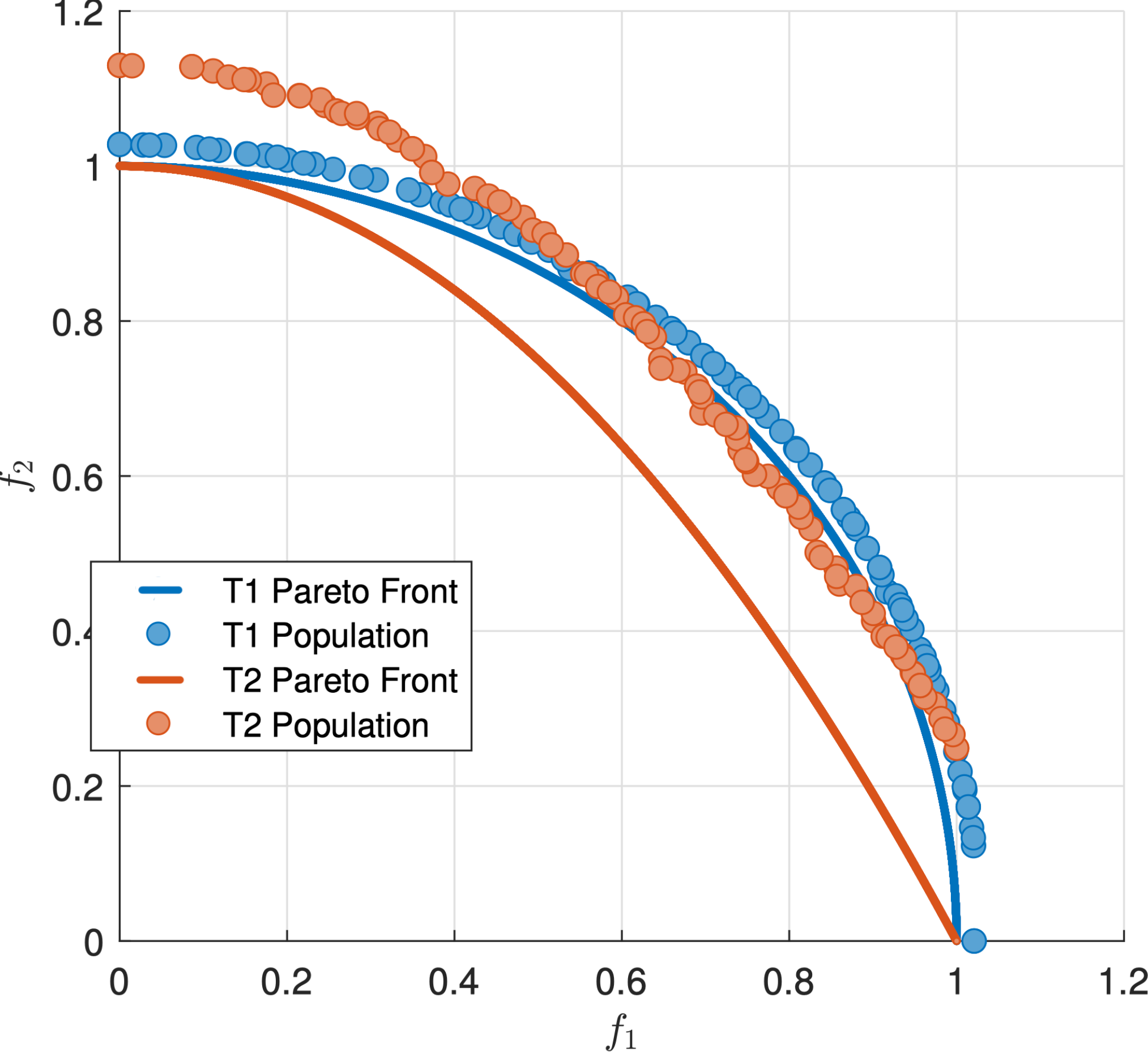}}
  \subfigure[Metric convergence]{\includegraphics[scale=0.43]{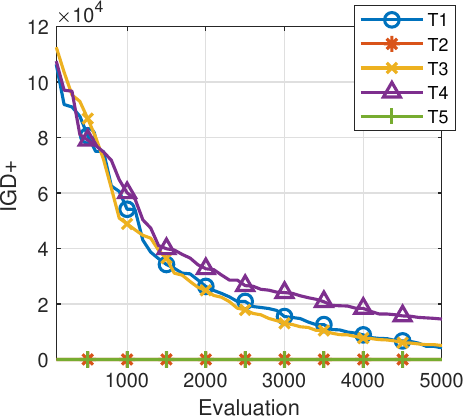}}
  \centering
  \caption{Examples of graphical display in the \code{Test Module} of MToP. (a) and (b) illustrate the landscapes of single-objective problems with different tasks in the one- and two-dimensional unified search space, respectively. (c) shows the feasible regions of a single-objective problem with different tasks in the two-dimensional unified search space. (d) depicts the Pareto front of a multi-objective problem with multiple tasks. (e) displays the convergence behavior of metrics after executing algorithms on problems.}
  \label{fig:test-module}
\end{figure*}

The \code{Test Module} is designed to assist researchers in the qualitative analysis of MTO problems and algorithms. It facilitates executing a single run of a selected algorithm on a chosen problem to generate a suite of diagnostic visualizations. These tools help inspect both problem characteristics and algorithmic performance, as illustrated in Fig.~\ref{fig:test-module}. For single-objective MTO problems, users can depict function landscapes in one or two dimensions (Fig.~\ref{fig:test-module}~(a)-(b)) or visualize the feasible regions of constrained tasks (Fig.~\ref{fig:test-module}~(c)). For multi-objective MTO problems, the module can plot the true Pareto fronts with populations (Fig.~\ref{fig:test-module}~(d)).

  {
    In addition to these static plots, the module provides dynamic visualization utilities. Once a run is initiated, researchers can enable options such as \code{Draw Dec} (decision space) and \code{Draw Obj} (objective space) to observe the population's evolution in real-time. Upon completion of the run, the final performance metrics are calculated, and their convergence behavior (as shown in Fig.~\ref{fig:test-module}~(e)) is displayed in a dedicated results panel on the right side of the interface.

    The \code{Test Module} serves as an essential preliminary step for researchers to understand the nuances of specific MTO problems and the behavior of algorithms before embarking on large-scale experiments.
  }




\subsubsection{Experiment Module}\label{sec:arch:func:exp}

\begin{figure}[htbp]
  \centering
  \subfigure[Convergence]{\includegraphics[scale=0.35]{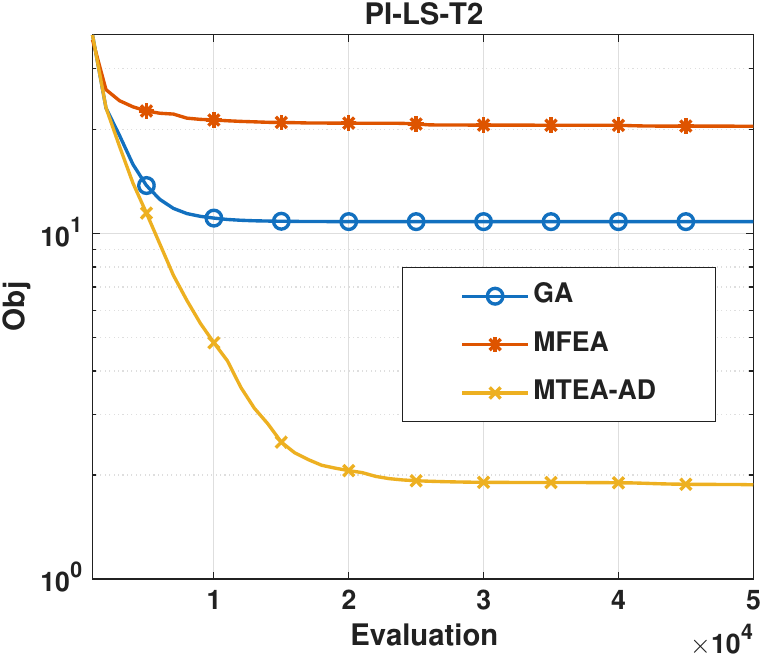}}
  \subfigure[Pareto front]{\includegraphics[scale=0.35]{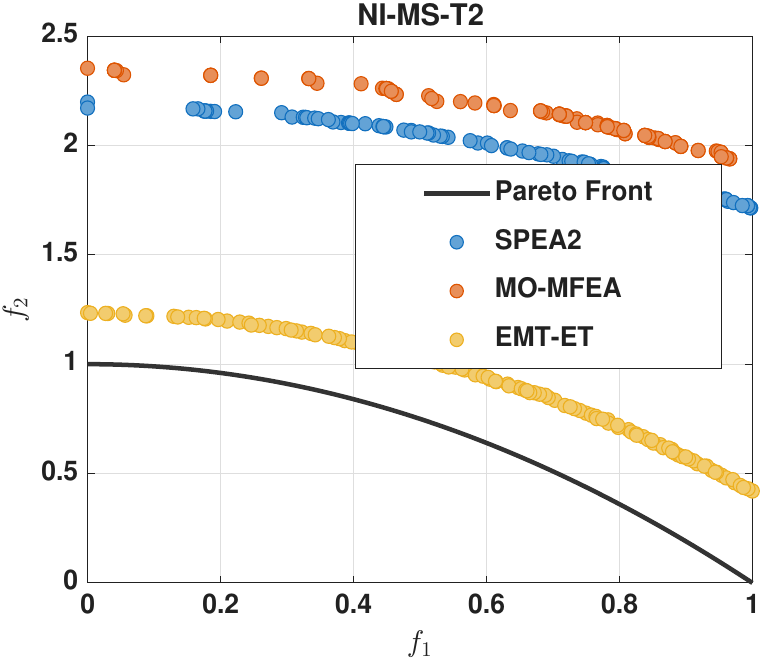}}
  \centering
  \caption{schematic plotting by the \code{Experiment Module} of MToP. (a) Convergence plot of algorithms on problems. (b) Pareto front plot of multi-objective problems.}
  \label{fig:exp-module}
\end{figure}

{
The \code{Experiment Module} provides the core functionalities for conducting comprehensive, large-scale experiments and analyzing the resulting data. It is designed to manage and execute batch runs, allowing users to test multiple algorithms across multiple MTO problems for a specified number of independent repetitions. Upon completion, the module systematically records all experimental settings, objective values, and decision variables. A key architectural feature is its support for MATLAB's Parallel Computing Toolbox, enabling the parallel execution of independent runs to significantly reduce total experimental time.

The module supports saving the complete, raw experimental dataset (\code{MTOData}) to a \code{.mat} file, which can be fully reloaded into the platform at any time. Once the data is generated, the module offers a suite of tools for post-processing and analysis. This includes functions to automatically calculate performance metrics, which are categorized to support both single-task and multi-task indicators, and to perform statistical significance tests to compare algorithmic performance. All generated tables and test results can be exported to \code{.xlsx}, \code{.csv}, or \code{.tex} formats. Furthermore, it provides granular export options for specific use cases, such as exporting convergence data in \code{.csv} format compatible with IOHanalyzer~\cite{Wang2022IOHanalyzer} or saving the optimal solution (single-objective MTO) or set of Pareto sets (multi-objective MTO) to \code{.mat} files for external study.

Finally, the module provides essential visualization capabilities for interpreting the experimental outcomes. As shown in Fig.~\ref{fig:exp-module}, this includes generating convergence plots to simultaneously compare the performance trajectories of multiple algorithms across various problems (Fig.~\ref{fig:exp-module}~(a)), a function applicable to both single- and multi-objective problems. Furthermore, for multi-objective MTO tasks, it can plot the acquired {Pareto front approximations} (Fig.~\ref{fig:exp-module}~(b)) to visually assess the quality of the final solution sets. Since all visualizations are generated as standard MATLAB figures, users can leverage MATLAB's built-in plotting tools for further customization, annotation, and export.
}

\subsubsection{Data Process Module}\label{sec:arch:func:data}

{
  Each unique execution within the \code{Experiment Module} is saved as a standardized data object to a \code{MTOData.mat} file, which is described in detail in Section~\ref{sec:arch:structure:data}, this file stores all experimental settings, results, and metadata. This saved data can be fully reloaded by the \code{Experiment Module} for post-processing or managed by the \code{Data Process Module}.

  The core function of the \code{Data Process Module} is to manage and manipulate these data objects, enabling data reuse and customization. It provides flexible \code{Merge} and \code{Split} capabilities. These operations can be performed along three distinct dimensions of the dataset: by independent runs (\eg, combining two 10-run sets into one 20-run set), by algorithms (\eg, adding a new algorithm's results to an existing dataset), or by problems (\eg, splitting a large benchmark suite into smaller subsets). This functionality is crucial for maintaining organized datasets and efficiently conducting comparative studies, such as when a new algorithm needs to be benchmarked against a set of previously executed experiments.
}

\subsection{Project Structure}\label{sec:arch:structure}

{
  The project structure of MToP is organized into three main perspectives: class diagram, sequence workflow, file structure, and experiment data structure. These perspectives collectively illustrate the static architecture, dynamic interactions, and organizational layout of MToP.
}

\subsubsection{Class Diagram}\label{sec:arch:structure:class}

\begin{figure}[htbp]
  \centering
  \includegraphics[scale=0.9]{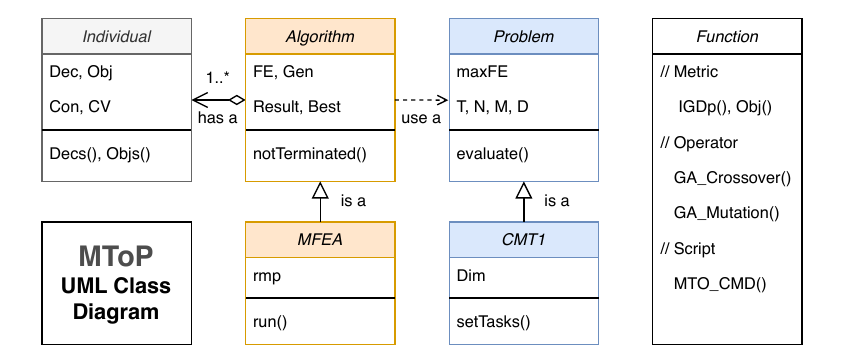}
  \centering
  \caption{Class diagram of MToP. The main classes include \code{Algorithm}, \code{Problem}, \code{Individual}.}
  \label{fig:class-diagram}
\end{figure}

{
Fig.~\ref{fig:class-diagram} illustrates the class diagram of MToP, which outlines the static architecture of the platform. The design is centered around three core base classes and functions:
\begin{itemize}
  \item \code{Individual} class is a data structure encapsulating a single solution, storing its decision variables (\code{Dec}), objective values (\code{Obj}), and constraint information (\code{Con}, \code{CV}).
  \item \code{Algorithm} class is the abstract base for all optimization algorithms with parameters ($FE$ for current function evaluations, $Gen$ for generations), managing the population of \code{Individual} objects (via an aggregation ``has a'' relationship) and the main evolutionary loop. A concrete example, \code{MFEA}, is shown inheriting from it. The \code{Algorithm} class also holds a ``use a'' dependency on the \code{Problem} class, as it needs to evaluate individuals.
  \item \code{Problem} class is an abstract base for optimization tasks, defining the essential \code{evaluate()} method and problem parameters ($T$ for task, $N$ for population size, $M$ for objective, $D$ for dimension).
  \item \code{Function} module groups static utility functions such as metrics (\eg, {\code{IGDp()}, \ie, \code{IGD+}}) and evolutionary operators (\eg, \code{GA\_Crossover()}).
\end{itemize}
}

\subsubsection{Sequence Workflow}\label{sec:arch:structure:workflow}

\begin{figure}[htbp]
  \centering
  \includegraphics[scale=0.6]{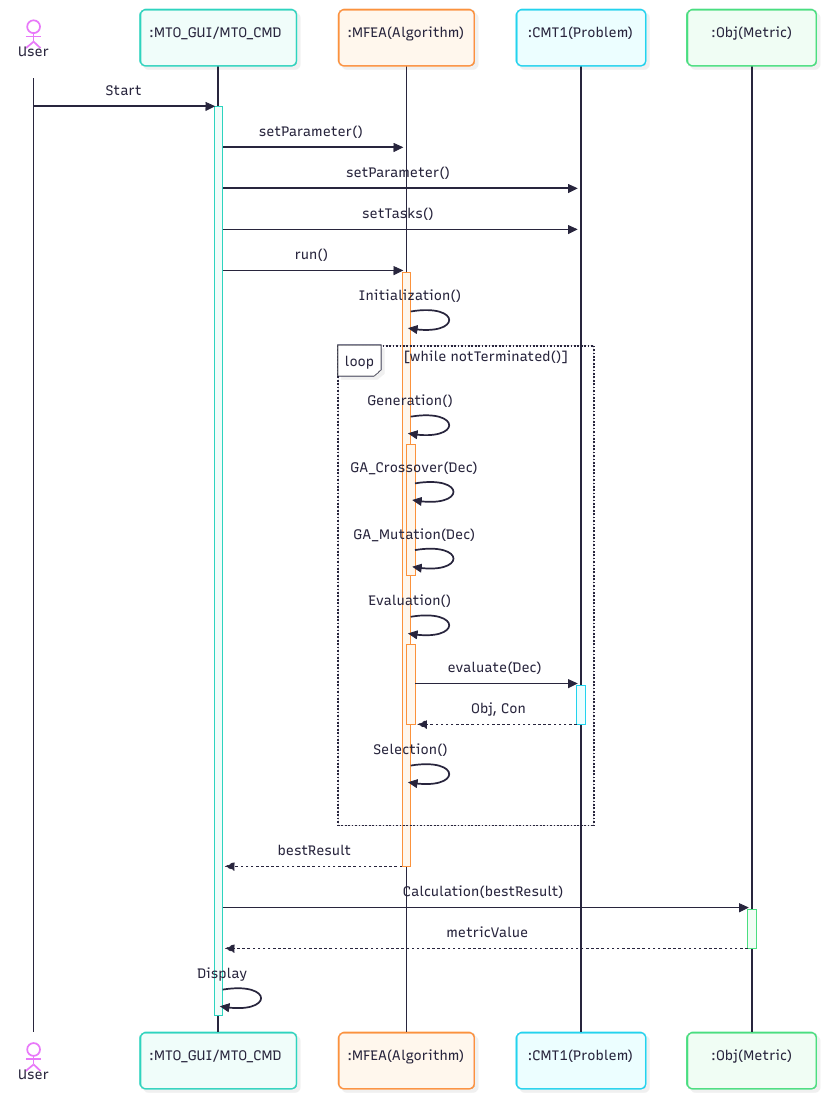}
  \centering
  \caption{Sequence diagram of MToP. The main workflow involves \code{MTO\_GUI} / \code{MTO\_CMD}, \code{Algorithm} {(\code{MFEA})}, \code{Problem} {(\code{CMT1})}, and \code{Metric} {(\code{Obj})}.}
  \label{fig:sequence-diagram}
\end{figure}

{
Fig.~\ref{fig:sequence-diagram} illustrates the dynamic interaction workflow using a concrete example of the \code{MFEA} algorithm solving the \code{CMT1} problem:
\begin{itemize}
  \item The process is initiated by a \code{User} through either the \code{MTO\_GUI} or \code{MTO\_CMD} interface.
  \item The interface configures the specific algorithm (\eg, \code{MFEA}) and problem (\eg, \code{CMT1}) instances by calling \code{setParameter()} and \code{setTasks()}. Subsequently, the interface invokes the \code{run()} method of \code{MFEA}, which performs \code{Initialization()} and enters the main optimization loop conditioned on \code{while notTerminated()}.
  \item Inside the loop, the workflow first executes the \code{Generation()} phase. Here, the algorithm calls specific operators, such as \code{GA\_Crossover()} and \code{GA\_Mutation()}, to produce offspring.
  \item Following generation, the \code{Evaluation()} method is triggered. This invokes the \code{evaluate()} method on the \code{CMT1} object, which returns the objective and constraint values (\code{Obj, Con}). Finally, the \code{Selection()} strategy is applied to determine the survivors for the next generation.
  \item Once the loop terminates, \code{MFEA} returns the \code{bestResult} to the interface. The interface then passes this result to the metric module (\eg, \code{Obj}) for \code{Calculation()}. Finally, the computed \code{metricValue} is displayed to the user.
\end{itemize}
}

\subsubsection{File Structure}\label{sec:arch:structure:structure}

\begin{figure}[htbp]
  \centering
  \includegraphics[scale=0.7]{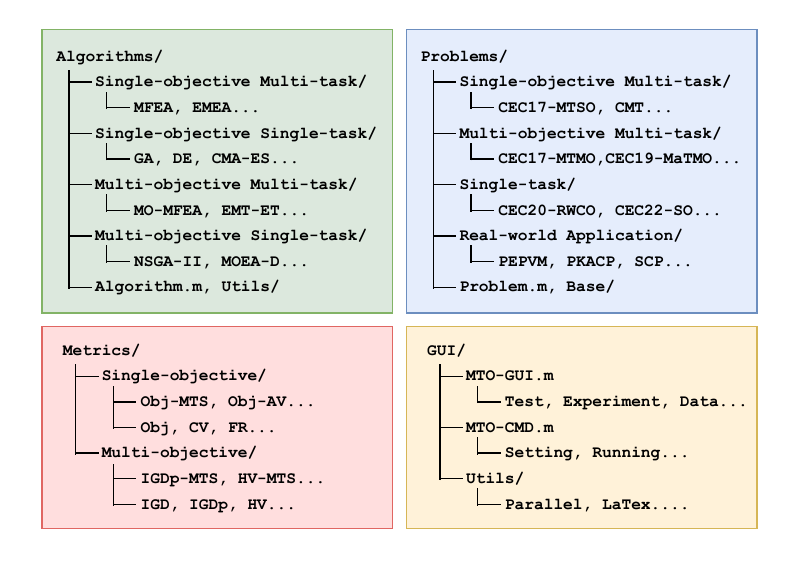}
  \centering
  \caption{File structure of MToP. The root directory contains the main script file \code{mto.m} and four subfolders: \code{Algorithms/}, \code{Problems/}, \code{Metrics/}, and \code{GUI/}.}
  \label{fig:file-structure}
\end{figure}

{ The file structure of MToP is shown in Fig.~\ref{fig:file-structure}. The root directory contains the main script file \code{mto.m} and four primary subfolders:}
\begin{itemize}
  \item \code{Algorithms/}: it contains various categories of algorithms, utility functions, and the algorithm base class \code{Algorithm.m}. Specific algorithm files such as \code{MFEA.m}, \code{MO\_MFEA.m}, and \code{GA.m} are stored within their respective classification folders. {The \code{Algorithms/Utils/} folder contains multiple subfolders, including \code{Operator/} and \code{Selection/}, which organize specific functional modules. The \code{Operator/} subfolder stores variation operators such as crossover and mutation, while the \code{Selection/} subfolder includes mating and environmental selection approaches. Other subfolders provide additional utility functions such as constraint handling and multi-objective optimization techniques.}
  \item \code{Problems/}: it {contains} various categories of problems, the base function folder, and the problem base class \code{Problem.m}. Specific problem files like \code{WCCI20\_MTSO1.m}, \code{CMT3.m}, and \code{CEC19\_MaTMO6.m} are stored under their corresponding classification folders. Real-world application problem files are located in the \code{Real-World Application/} folder, while the \code{Problems/Base/} folder contains public base functions for problems.
  \item \code{Metrics/}: it contains all result evaluation metrics, organized into subfolders for single-objective \code{Metrics/Single-objective/} and multi-objective \code{Metrics/Multi-objective} optimization. Specific metric files include \code{Obj.m}, \code{Obj\_MTS.m}, and \code{IGDp.m}, among others.
  \item \code{GUI/}: it {contains} all files used by the GUI of MToP. Among them, \code{MTO\_GUI.m} serves as the main file of the GUI, while \code{MTO\_CMD.m} provides functionality for executing experiments via the command line.
\end{itemize}

\subsubsection{Experiment Data Structure}\label{sec:arch:structure:data}

{
  The experiment data structure is the standardized format used by MToP to save the complete output of an experimental batch, regardless of whether it is generated by the \code{MTO\_CMD.m} command-line runner or the \code{MTO\_GUI.m} graphical interface. All experimental data is encapsulated within a single MATLAB struct variable named \code{MTOData}, which is then saved to a \code{.mat} file (\eg, \code{MTOData.mat}). This self-contained design ensures portability and simplifies downstream data management and post-processing.
}

\begin{table}[htbp]
  \centering
  \renewcommand{\arraystretch}{1.2}
  \caption{Properties of experimental data.}
  \resizebox{.6\textwidth}{!}{
    \begin{tabular}{ll}
      \toprule
      Property          & Description                                                       \\
      \midrule
      \code{Reps}       & Repetitions number of independent runs                            \\
      \code{Algorithms} & Algorithms data contains names and parameter settings             \\
      \code{Problems}   & Problems data contains names and parameter settings               \\
      \code{Results}    & Total results data contains \code{Obj}, \code{CV}, and \code{Dec} \\
      \code{RunTimes}   & Running time data of algorithms on problems                       \\
      \code{Metrics}    & Calculated metric results data                                    \\
      \bottomrule
    \end{tabular}}
  \label{tab:prop-data}
\end{table}

{
The high-level properties stored within the \code{MTOData} struct are summarized in Table~\ref{tab:prop-data}. The core fields are organized as multi-dimensional arrays to map clearly onto the experimental design. Let $P$ be the number of problems, $A$ be the number of algorithms, and $R$ be the number of \code{Reps}.
\begin{itemize}
  \item \code{Problems}: This is a $1 \times P$ struct array. Each element \code{MTOData.Problems(i)} stores the static metadata for the $i$-th problem, including its \code{Name}, number of tasks (\code{T}), objective dimensions (\code{M}), decision variable dimensions (\code{D}), and maximum function evaluations (\code{maxFE}).
  \item \code{Algorithms}: This is a $1 \times A$ struct array. Each element \code{MTOData.Algorithms(j)} stores the metadata for the $j$-th algorithm, such as its \code{Name} and a struct of its specific parameters (\code{Para}).
  \item \code{RunTimes}: This is a $P \times A \times R$ numerical matrix, where the element \code{(i, j, k)} stores the wall-clock execution time for running algorithm $j$ on problem $i$ during repetition $k$.
  \item \code{Results}: This is the $P \times A \times R$ struct array containing the core optimization output. Each struct \code{MTOData.Results(i, j, k)} contains the detailed results for a single run, aggregated across all $T$ tasks and all $G$ saved {checkpoints} (where $G$ is the \code{Results Num}). Each struct contains:
        \begin{itemize}
          \item \code{Obj}: Stores the objective values. For single-objective problems, this is a $T \times G$ numerical array storing the single best value for each task at each {checkpoint}. For multi-objective problems, this is a $T \times 1$ cell array, where each cell \code{Obj\{t\}} contains a $G \times N \times M_t$ array representing the full objective values of $N$ solutions.
          \item \code{CV}: Stores the constraint violation values. This is typically a $T \times G$ array (for single-objective problems) or a $T \times G \times N$ array (for multi-objective problems) storing the summed constraint violation for the corresponding solutions in \code{Obj}.
          \item \code{Dec}: Stores the decision variables, which are populated only if the \code{Save\_Dec} flag is set to true. For single-objective problems, this is a $T \times G \times D_t$ array storing the decision vector of the best solution. For multi-objective problems, it is a $T \times G \times N \times D_t$ array storing the decision vectors for all $N$ solutions.
        \end{itemize}
  \item \code{Metrics}: This field is not generated by the experiment runners (CMD or GUI) themselves but is designed to be populated later by the \code{Experiment Module} when metrics are calculated and saved back to the data file.
\end{itemize}
This consistent, multi-dimensional struct-based layout is fundamental to the platform, as it allows the \code{Data Process Module} to reliably merge and split datasets by manipulating the \code{Problems}, \code{Algorithms}, or \code{Reps} dimensions of this structure.
}

\subsection{Code Pattern}\label{sec:arch:code}

{
  This subsection details the code patterns employed in MToP for implementing algorithms, problems, and metrics. By adhering to these standardized patterns, developers can easily extend the platform with new methods while ensuring compatibility with the existing architecture and GUI.
}

\subsubsection{Algorithm}\label{sec:arch:code:algo}

\begin{table}[htbp]
  \centering
  \renewcommand{\arraystretch}{1.2}
  \caption{Properties and methods of algorithm base class.}
  \resizebox{.65\textwidth}{!}{
    \begin{tabular}{ll}
      \toprule
      Property or method     & Description                                                \\
      \midrule
      \code{FE}              & Number of fitness function evaluations                     \\
      \code{Gen}             & Number of evolutionary generations                         \\
      \code{Best}            & Best individual found for single-objective optimization    \\
      \code{Result}          & Result data contains \code{Obj}, \code{CV}, and \code{Dec} \\
      \code{getParameter()}  & Get customized parameters for algorithm object             \\
      \code{setParameter()}  & Set customized parameters for algorithm object             \\
      \code{notTerminated()} & Determine whether to terminate and update result data      \\
      \code{Evaluation()}    & {Function evaluation with algorithm state update}          \\
      \code{run()}           & Executing algorithm                                        \\
      \bottomrule
    \end{tabular}}
  \label{tab:prop-method-algo}
\end{table}

\begin{lstlisting} [
    label={list:algo},
    caption={Algorithm implementation example for multi-population MTEA.},
    captionpos=b,
    float=htbp
]
classdef Algo_Example1_MP < Algorithm
% <Multi-task> <Single-objective> <None/Constrained>

properties % set parameter
  para_example = 0.5;
end

methods
function run(Algo, Prob)
  % initialize multiple populations for all tasks
  pops = Initialization(Algo, Prob, Individual);
  % main loop
  while Algo.notTerminated(Prob, pops) 
    for t = 1:Prob.T % for each task
      % generate new offspring
      offspring(index1) = Generation(pops{t});
      % knowledge transfer
      offspring(index2) = KnowledgeTransfer(pops);
      % evaluate the fitness of offspring
      offspring = Algo.Evaluation(offspring, Prob, t);
      % environmental selection
      pops{t} = Selection(pops{t}, offspring);
    end
  end
end
\end{lstlisting}

\begin{lstlisting} [
    label={list:algo2},
    caption={Algorithm implementation example for multifactorial MTEA.},
    captionpos=b,
    float=htbp
]
classdef Algo_Example2_MF < Algorithm
% <Multi-task> <Multi-objective> <None/Constrained>

properties % set parameter
  para_example = 0.2;
end

methods
function run(Algo, Prob)
  % initialize populations for multifactorial evolution
  population = Initialization_MF(Algo, Prob, Individual);
  % main loop
  while Algo.notTerminated(Prob, population)
    % generate new offspring with random mating
    offspring = Generation(population);
    % evaluate based on offspring skill factor
    for t = 1:Prob.T
      idx = [offspring.MFFactor] == t;
      offspring(idx) = Algo.Evaluation(offspring(idx), Prob, t);
    end
    % environmental selection
    population = Selection(population, offspring);
  end
end
\end{lstlisting}

All algorithms within MToP inherit from the algorithm base class \code{Algorithm.m}. The properties and methods of this base class are detailed in Table~\ref{tab:prop-method-algo}. This base class encapsulates all functions that interface with the GUI, simplifying the implementation process for algorithms, which only need to focus on the evolutionary workflow itself. The MTEAs implemented in MToP are outlined in Table~\ref{tab:algo-mt} {of the supplementary files.} Additionally, the single-task EAs are exclusively implemented using multi-population methods tailored for MTO problems. Specific algorithms belonging to this category are listed in Table~\ref{tab:algo-st}.

Listings~\ref{list:algo} and \ref{list:algo2} exemplify the implementation of multi-population and multifactorial algorithms respectively. The algorithm labels provided in the second line serve as identifiers for classification within the GUI. The \code{run()} function within each class orchestrates algorithm execution, with \code{Algo} representing the object itself and \code{Prob} denoting the problem object to be solved. Population initialization occurs at the outset of the \code{run()} function, utilizing either the multi-population or multifactorial method provided by MToP. {Here, most standard algorithms read the \code{N} property from the \code{Prob} object (see Table~\ref{tab:prop-method-prob}) to use the problem's default population size, which ensures consistency for comparisons. However, this is not a rigid requirement: algorithms known for dynamic population sizing (\eg, L-SHADE, IPOP-CMA-ES) are implemented to override this default and manage their own population sizes as per their original designs.}

  {
    In Listing~\ref{list:algo}, the variable \code{pops} is a \code{cell array}. It stores multiple populations grouped by task, in the form of \code{\{pop1, pop2, pop3, ...\}}. Each element \code{pop} is itself a separate \code{object array} containing all individuals \code{[ind1, ind2, ind3, ...]} for a specific task. Each \code{Individual} object within these lists contains properties such as \code{Obj} (objective value), \code{Con} (constraint value), and \code{Dec} (decision variables).
    In contrast, Listing~\ref{list:algo2} uses a \code{object array} data type for the whole \code{population} variable. This adopts a mixed structure, storing individuals from all tasks together in a single list. To differentiate which task each individual belongs to, the \code{Individual} objects in this listing include an additional property, \code{MFFactor}. Therefore, the structure of this single list can be conceptualized as \code{[ind1, ind2, ind3, ...]}, where each individual's \code{MFFactor} property is set to its corresponding task ID (\eg, \code{ind1.MFFactor = 1, ind2.MFFactor = 2, ind3.MFFactor = 1, ...}).
  }

Subsequently, the primary loop commences with the invocation of the \code{notTerminated()} function, a component of the algorithm base class. This function oversees data updates and generation counting within the loop. During the main loop, distinct operations are carried out for offspring generation \code{Generation()}, offspring evaluation \code{Evaluation()}, and environmental selection \code{Selection()}.

In the context of knowledge transfer, the multi-population algorithm requires the implementation of the \code{KnowledgeTransfer()} function to acquire knowledge from other tasks. Conversely, the multifactorial approach achieves knowledge transfer through random mating within the \code{Generation()} function. The \code{Generation()} function operates on the decision variables \code{Dec} of offspring individuals and is tailored to specific algorithms.

  {
    The \code{Evaluation()} method serves as a state-updating wrapper in \code{Algo}. It performs two sequential operations: first, it invokes the \code{evaluate()} method from the \code{Prob} object to calculate the actual \code{Obj} and \code{Con} values. Second, it updates the algorithm's internal state properties, such as incrementing the \code{FE} counter.
  }

Following, the environmental selection function \code{Selection()} is invoked to update the new population. While MToP offers universal \code{Selection()} functions, specific algorithms also have the option to reimplement this function. Subsequently, the code progresses to the next loop and invokes \code{notTerminated()} to document changes. The algorithmic structure in MToP is designed to accommodate all EAs for solving MTO problems.

\subsubsection{Problem}\label{sec:arch:code:prob}

All problems within MToP inherit from the problem base class \code{Problem.m}. The properties and methods of \code{Problem.m} are detailed in Table~\ref{tab:prop-method-prob}. As dimensions and upper and lower boundaries may vary across tasks, MToP offers a default unified search space mapping approach as Eq.~\eqref{eq:uni-space}. It's important to note that while the unified search space approach serves as the default mapping method in MToP, alternative mapping techniques can also be implemented within specific algorithms.

A simple problem implementation example is illustrated in Listing~\ref{list:prob}. The maximum number of function evaluations, denoted as \code{maxFE}, for the problem can be specified within the class constructor. The function \code{setTasks()} is responsible for configuring the properties and evaluation function for each optimization task. Within this function, the number of tasks \code{T} is first defined, followed by a detailed setup for each task. Subsequently, parameters such as decision variable dimensions \code{D}, objective dimensions \code{M}, fitness function \code{Fnc}, lower bound \code{Lb}, and upper bound \code{Ub} are set individually for each task. For problems involving multiple tasks, \code{M} and \code{D} are represented as \code{vector} type, while \code{Fnc}, \code{Lb}, and \code{Ub} are represented as \code{cell} type.

  {
    Notably, the function \code{Fnc} for each task takes the decision variable \code{Dec} as input and returns objective values \code{Obj} along with the constraint value \code{Con}.
    The problem's base class \code{evaluate()} method manages this process by using the specific function handle defined in \code{Fnc} to perform the calculation for the requested task.
  }

  {
    Moreover, for multi-objective optimization, the external function \code{getOptimum()} is provided. It defines the true optimal solutions (\ie, true Pareto front) if known, which is required for metrics like \code{IGD+}. For problems where the true front is unknown (\eg, real-world applications), this function is instead used to provide the reference point for calculating metrics such as hypervolume (HV).
  }

With these standardized problem code patterns, all fully defined problems can be solved using the corresponding types of algorithms in MToP. The existing problems available in MToP are enumerated in Table~\ref{tab:prob-mt} {of the supplementary files}.

\begin{table}[htbp]
  \centering
  \renewcommand{\arraystretch}{1.2}
  \caption{Properties and methods of problem base class.}
  \resizebox{.6\textwidth}{!}{
    \begin{tabular}{ll}
      \toprule
      Property or method       & Description                                            \\
      \midrule
      \code{T}                 & Number of optimization tasks                           \\
      \code{N}                 & {Default population size} for each task                \\
      \code{M}                 & Number of objectives for all tasks                     \\
      \code{D}                 & Number of decision variable dimensions for all tasks   \\
      \code{Fnc}               & Fitness function for all tasks                         \\
      \code{Lb}                & Lower bound of decision variables for all tasks        \\
      \code{Ub}                & Upper bound of decision variables for all tasks        \\
      \code{maxFE}             & Maximum fitness function evaluations                   \\
      \code{getRunParameter()} & Get public parameters for problem object               \\
      \code{getParameter()}    & Get customized parameters for problem object           \\
      \code{setParameter()}    & Set customized parameters for problem object           \\
      \code{setTasks()}        & Set optimization tasks                                 \\
      \code{getOptimum()}      & Get optimal solutions for multi-objective optimization \\
      \code{evaluate()}        & Fitness function evaluation                            \\
      \bottomrule
    \end{tabular}}
  \label{tab:prop-method-prob}
\end{table}

\begin{lstlisting} [
    label={list:prob},
    caption={Problem implementation example.},
    captionpos=b,
    float=htbp
]
classdef Prob_Example < Problem
% <Multi-task> <Multi-objective> <None>

methods
function Prob = Prob_Example(varargin)
  Prob = Prob@Problem(varargin);
  % set default maximum function evaluations
  Prob.maxFE = 1000 * 100;
end
function setTasks(Prob)
  Prob.T = 2;
  % task 1
  Prob.D(1) = 10; % variable dimensions
  Prob.M(1) = 2; % objective number
  Prob.Fnc{1} = @func1 % fitness function
  Prob.Lb{1} = zeros(1, 10); % lower bound
  Prob.Ub{1} = ones(1, 10); % upper bound
  % task 2
  Prob.D(2) = 20; % variable dimensions
  Prob.M(2) = 3; % objective dimensions
  Prob.Fnc{2} = @func2 % objective function
  Prob.Lb{2} = [0,-ones(1, 19)]; % lower bound
  Prob.Ub{2} = [1,ones(1, 19)]; % upper bound
end
function optimum = getOptimum(Prob) % optional
  % return optimum points for each task
  optimum{1} = optimum_matrix1;
  optimum{2} = optimum_matrix2;
end
end
\end{lstlisting}

\subsubsection{Performance Metric}\label{sec:arch:code:metric}

\begin{lstlisting} [
    label={list:metric},
    caption={Metric implementation example.},
    captionpos=b,
    float=htbp
]
function result = Metric_Example(MTOData)
% <Metric> <Multi-objective>

  result.Metric = 'Min';

  % Data for shown in the GUI table
  result.RowName = {MTOData.Problems.Name};
  result.ColumnName = {MTOData.Algorithms.Name};
  result.TableData = CalculateTableData(MTOData);

  % Data for shown in the GUI convergence plot (optional)
  result.ConvergeData = CalculateConvergeData(MTOData);

  % Data for shown in the GUI Pareto plot (optional)
  result.ParetoData = CalculateParetoData(MTOData);
end
\end{lstlisting}

Unlike the implementation of algorithm and problem classes, metric codes in MToP are defined as functions. An illustrative example of metric implementation is presented in Listing~\ref{list:metric}. The input parameter of the function is \code{MTOData}, which is generated during experimental execution. The function returns \code{result}, comprising properties such as \code{Metric}, \code{RowName}, \code{ColumnName}, \code{TableData}, \code{ConvergeData}, and \code{ParetoData}. {The \code{Metric} property can take values of either \code{Min} or \code{Max}, indicating whether a smaller or larger metric value is preferable.} \code{RowName}, \code{ColumnName}, and \code{TableData} are utilized to present metric results in the GUI table. On the other hand, \code{ConvergeData} and \code{ParetoData} are employed to exhibit metric convergence results for convergence plots and non-dominated solutions for Pareto plots, respectively. Note that specific metric calculation functions are not elaborated here for the sake of simplicity. The metrics currently integrated into MToP are enumerated in {Table~\ref{tab:metric}}.

\begin{table}[htbp]
  \centering
  \renewcommand{\arraystretch}{1.2}
  \caption{Metrics in MToP. Abbreviations: SO/MO = single-/multi-objective; ST/MT = single-/multi-task; CV = constraint violation; FR = feasible rate; HV = hypervolume; IGD = inverted generational distance; IGD+ = improved IGD plus; AV/UV = average/unified average; MTS = multitask score; CMT = competitive multitask; NBR = number of best results.}
  \resizebox{.55\textwidth}{!}{
    \begin{tabular}{lccl}
      \toprule
      Metric       & Objective & Task  & Description                                  \\
      \midrule
      Obj          & SO        & ST/MT & Objective value for each task                \\
      Obj (AV)     & SO        & MT    & Average Obj for all tasks                    \\
      Obj (UV)     & SO        & MT    & Unified average Obj for all tasks            \\
      Obj (MTS)    & SO        & MT    & Multitask score of Obj for all tasks         \\
      Obj (CMT)    & SO        & MT    & Competitive multitask Obj for all tasks      \\
      Obj (NBR)    & SO        & MT    & Number of best Obj result for all tasks      \\
      CV           & SO        & ST/MT & Constraint violation for each task           \\
      FR           & SO        & ST/MT & Feasible rate for each task                  \\
      HV           & MO        & ST/MT & Hypervolume for each task                    \\
      HV (MTS)     & MO        & MT    & Multitask score of HV for all tasks          \\
      HV (CMT)     & MO        & MT    & Competitive multitask HV for all tasks       \\
      IGD          & MO        & ST/MT & Inverted generational distance for each task \\
      IGD (AV)     & MO        & MT    & Average IGD for all tasks                    \\
      IGD (MTS)    & MO        & MT    & Multitask score of IGD for all tasks         \\
      IGD (CMT)    & MO        & MT    & Competitive multitask IGD for all tasks      \\
      IGD+         & MO        & ST/MT & Improved IGD plus for each task              \\
      IGD+ (MTS)   & MO        & MT    & Multitask score of IGD+ for all tasks        \\
      IGD+ (CMT)   & MO        & MT    & Competitive multitask IGD+ for all tasks     \\
      Spread       & MO        & ST/MT & Spread metric for each task                  \\
      Spread (CMT) & MO        & MT    & Competitive multitask spread for all tasks   \\
      Run Time     & SO/MO     & ST/MT & Algorithm running time for all tasks         \\
      \bottomrule
    \end{tabular}}
  \label{tab:metric}
\end{table}

\section{Guidelines for Using MToP}\label{sec:guide}

\begin{figure*}[htbp]
  \centering
  \subfigure[Test Module]{\includegraphics[scale=0.24]{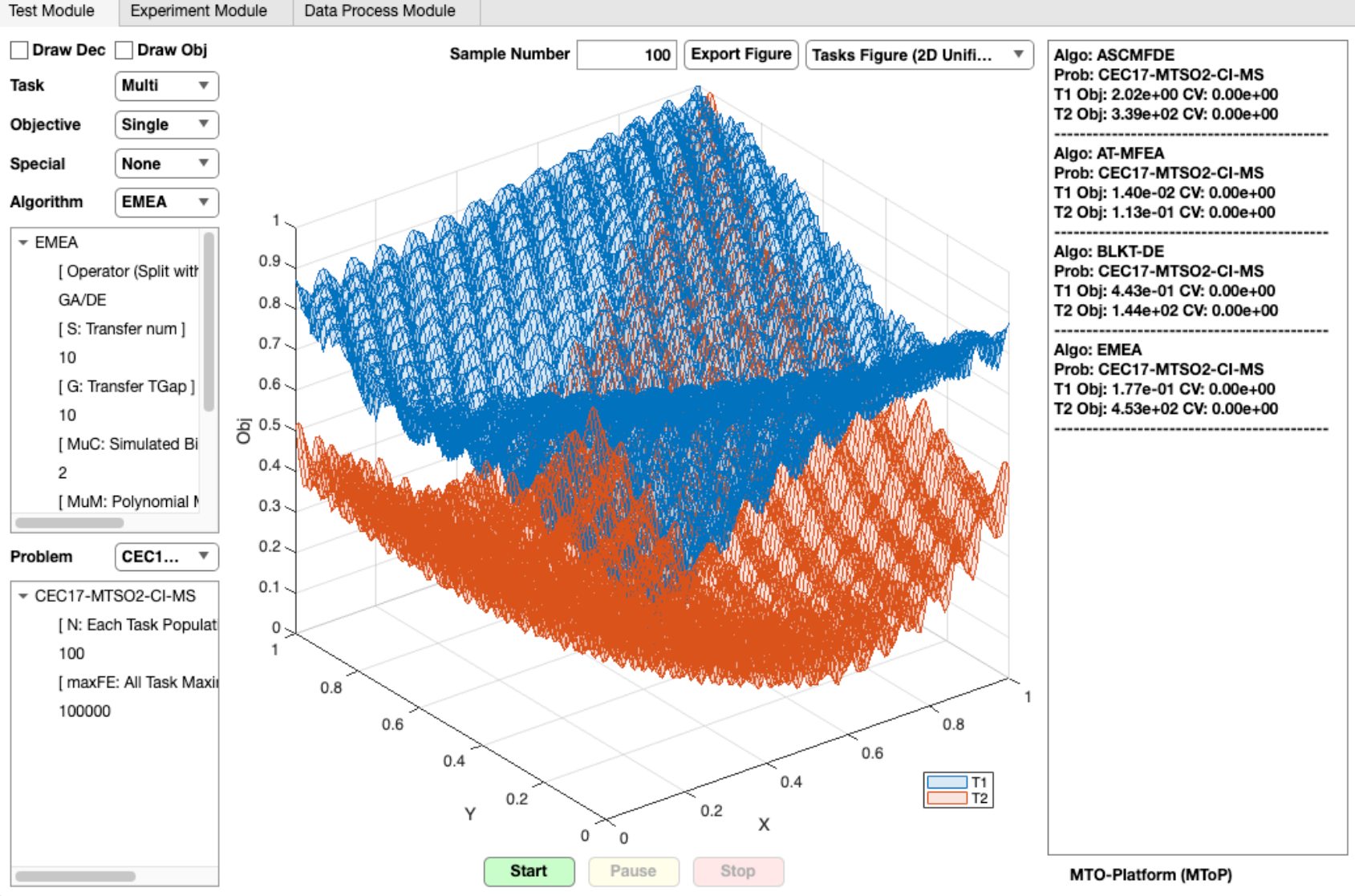}}
  \subfigure[Experiment Module]{\includegraphics[scale=0.24]{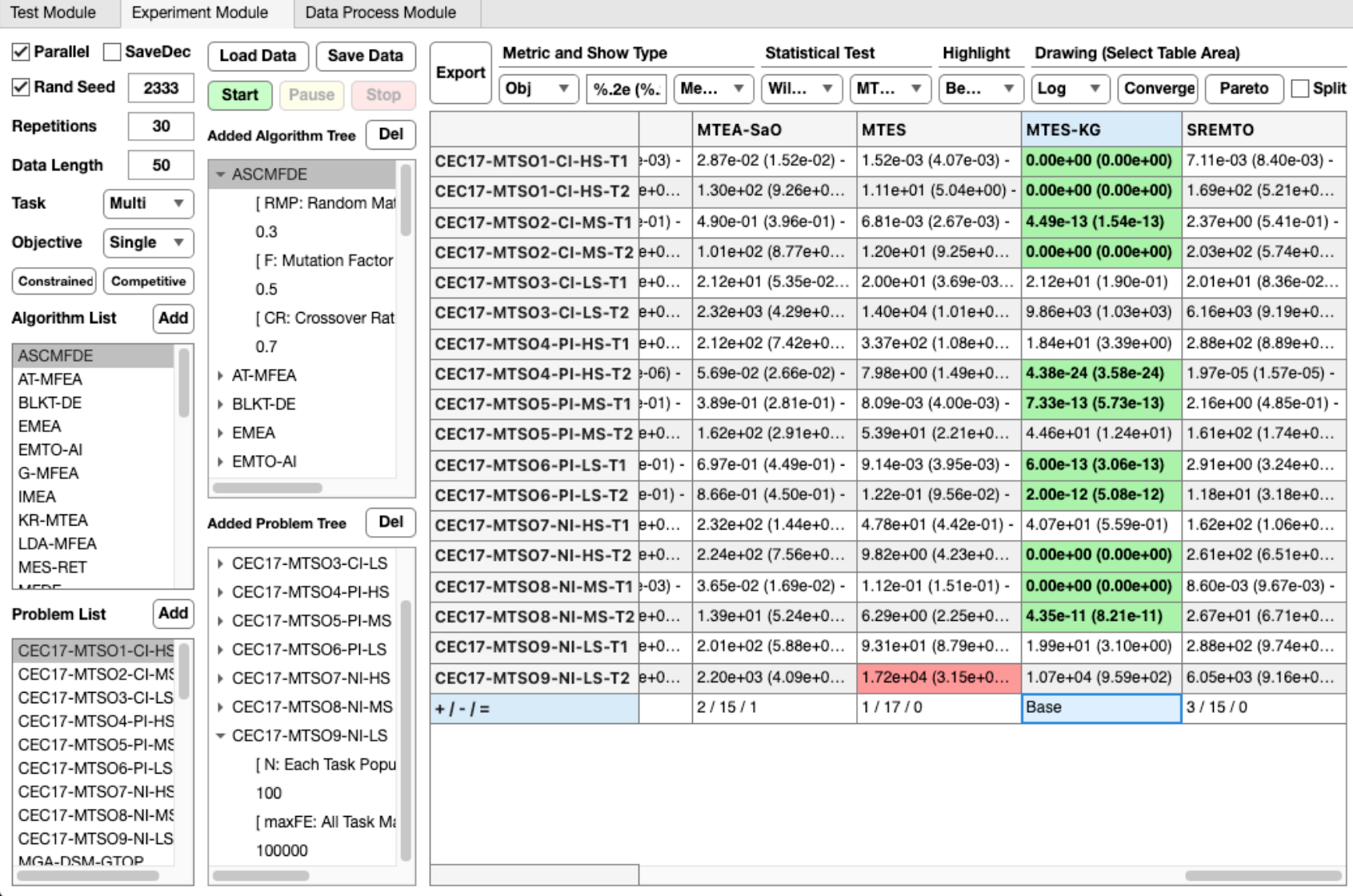}}
  \centering
  \caption{Graphical user interface of MToP. The \code{Test Module} (a) is used for testing algorithms and problems. The graphical display is shown in the center section of the \code{Test Module}. The \code{Experiment Module} (b) is used for executing comparative experiments. The results of the experiment are displayed in the right section of the \code{Experiment Module}.}
  \label{fig:mtop-module}
\end{figure*}

To launch MToP, start by running the script file \code{mto.m} located in the root directory. This action will initialize the MToP GUI interface, which is illustrated in Fig.~\ref{fig:mtop-module}. The GUI interface of MToP requires MATLAB R2022b or later versions to run, while command-line execution can be done with any version.

\subsection{Testing Algorithms and Problems}\label{sec:guide:test}

As depicted in Fig.~\ref{fig:mtop-module}~(a), the \code{Test Module} interface is structured into left, center, and right sections. The left section is dedicated to configuration, the center section serves as the main display for visualizations, and the right section shows the final metric results after a run. This module is designed for preliminary analysis, debugging, and qualitative visualization of a single algorithm's performance on a single problem.

\subsubsection{Viewing Problem Characteristics}\label{sec:guide:test:prob}

\begin{figure*}[htbp]
  \centering
  \includegraphics[scale=0.3]{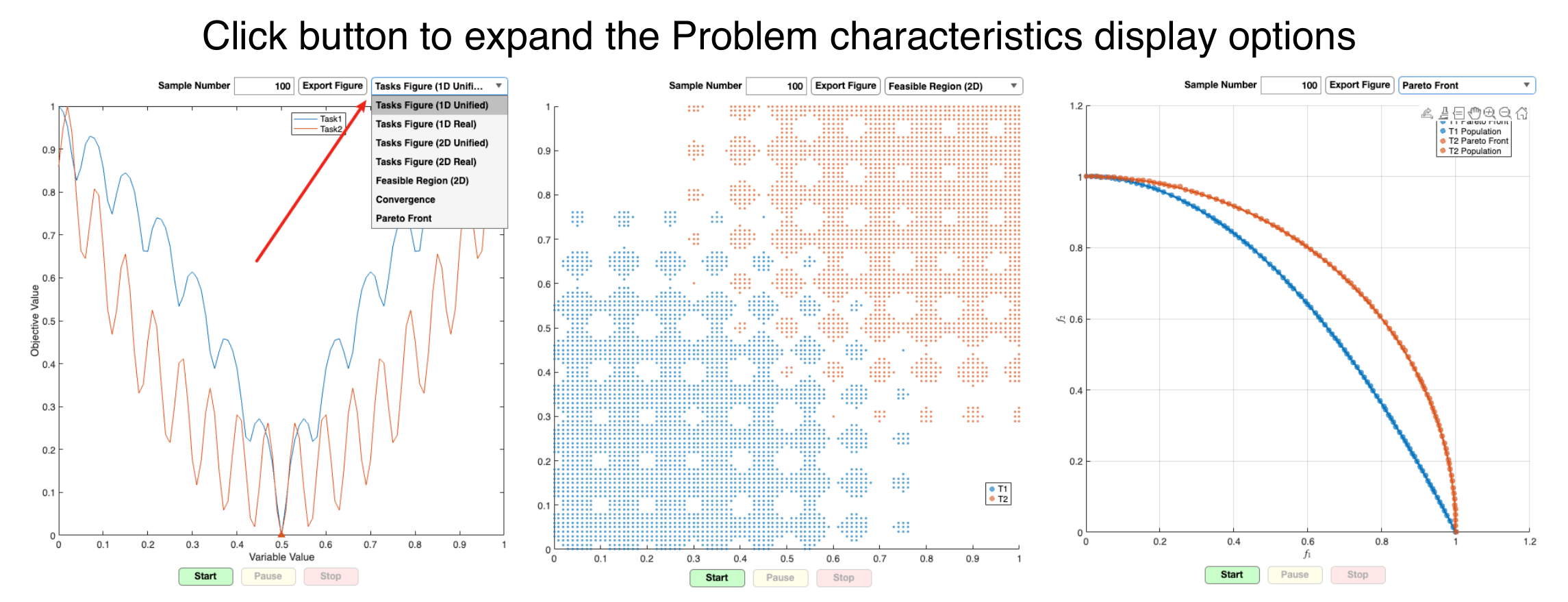}
  \centering
  \caption{Problem characteristics visualization in the \code{Test Module} of MToP GUI.}
  \label{fig:test-prob}
\end{figure*}

{
Before executing an algorithm, researchers can use the \code{Test Module} to visualize the characteristics of the selected problem. As shown in Fig.~\ref{fig:test-prob}, after selecting a problem (\eg, CEC17-MTMO1-CI-HS), the central display area provides a drop-down menu (initially labeled \code{Tasks Figure (1D Unified)} in the figure). Clicking this menu reveals various visualization options. These options vary by problem type but include plotting 1D or 2D function landscapes to understand the search space, visualizing feasible regions for constrained problems, and displaying the true Pareto front for multi-objective tasks after running the algorithm. This allows for an initial assessment of the problem's difficulty and features prior to optimization.
}

\subsubsection{Setting Algorithm and Problem Parameters}\label{sec:guide:test:param}

\begin{figure*}[htbp]
  \centering
  \includegraphics[scale=0.46]{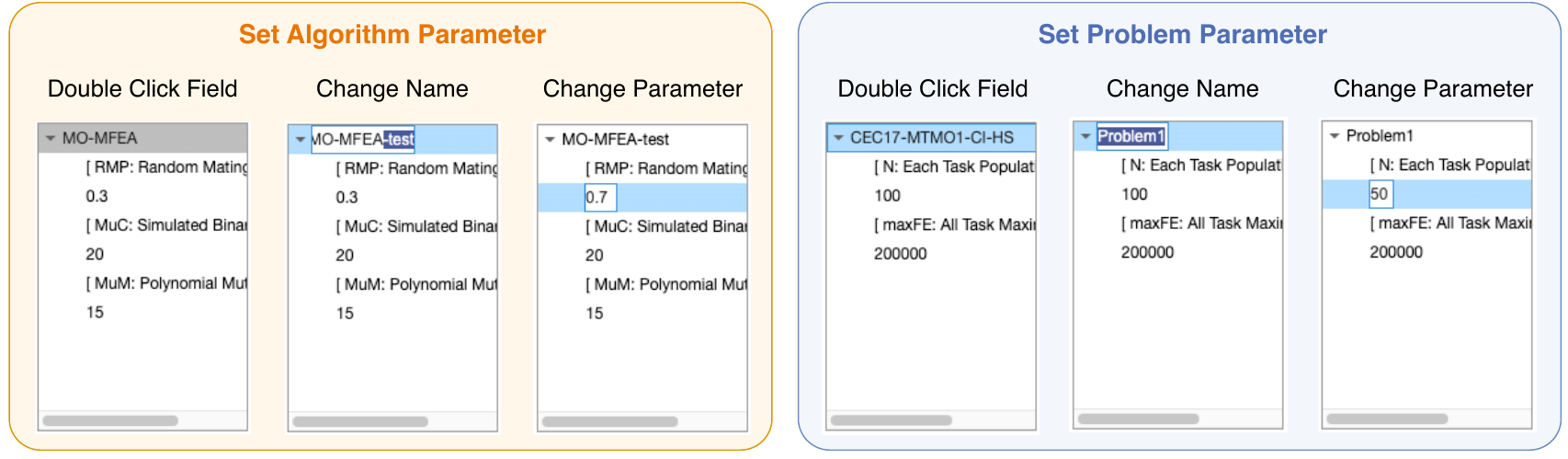}
  \centering
  \caption{Algorithm and problem parameter setting in the \code{Test Module} of MToP GUI.}
  \label{fig:test-para}
\end{figure*}

{
The left section of the module handles the selection and configuration of algorithms and problems. As illustrated in Fig.~\ref{fig:test-para}, users first select the desired algorithm (\eg, MO-MFEA) and problem (\eg, CEC17-MTMO1-CI-HS) from the respective lists. Filtering options (\eg, \code{Task}, \code{Objective}, \code{Special}) help narrow down the choices. Once selected, their default parameters appear in the text boxes below. To modify a parameter, the user can double-click the parameter's name in the list, which opens an editable field (as shown in the figure for \code{Change Name} and \code{Change Parameter}). For example, the algorithm's name can be changed, or a specific numeric parameter like the problem's population size \code{N} can be adjusted. These changes are applied to the objects when the experiment is started by pressing the \code{Start} button in the central panel.
}

\subsubsection{Visualizing Algorithm Behavior}\label{sec:guide:visual}

\begin{figure*}[htbp]
  \centering
  \subfigure[Gen=1]{\includegraphics[scale=0.2]{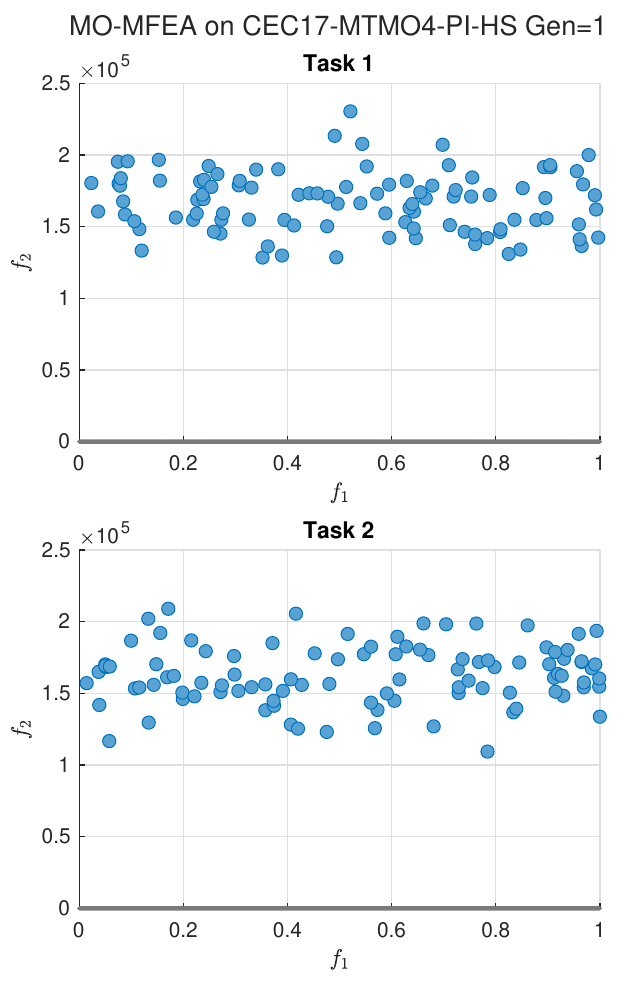}}
  \subfigure[Gen=200]{\includegraphics[scale=0.2]{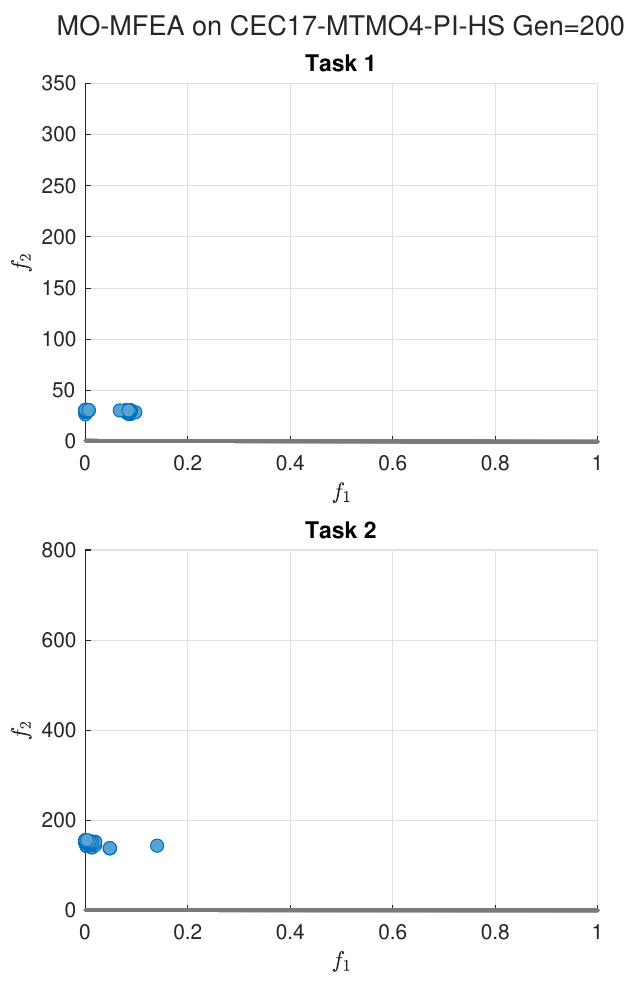}}
  \subfigure[Gen=500]{\includegraphics[scale=0.2]{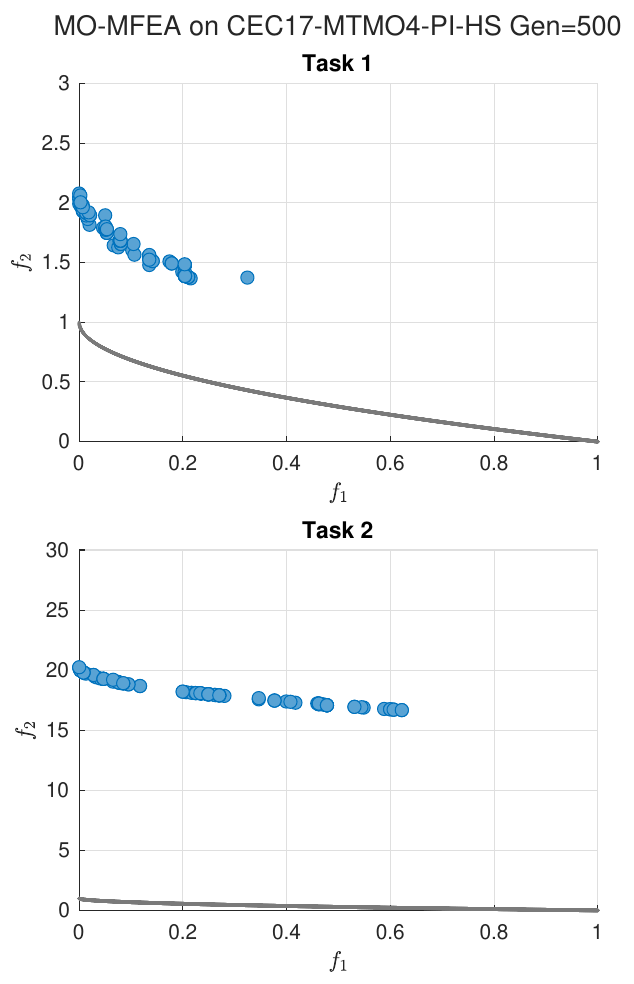}}
  \subfigure[Gen=700]{\includegraphics[scale=0.2]{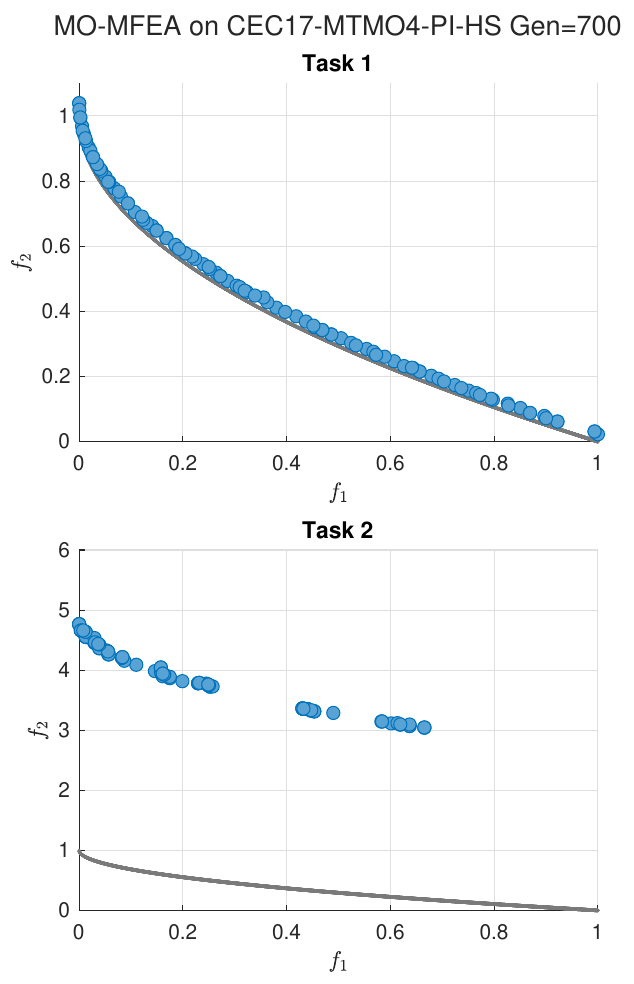}}
  \subfigure[Gen=1000]{\includegraphics[scale=0.2]{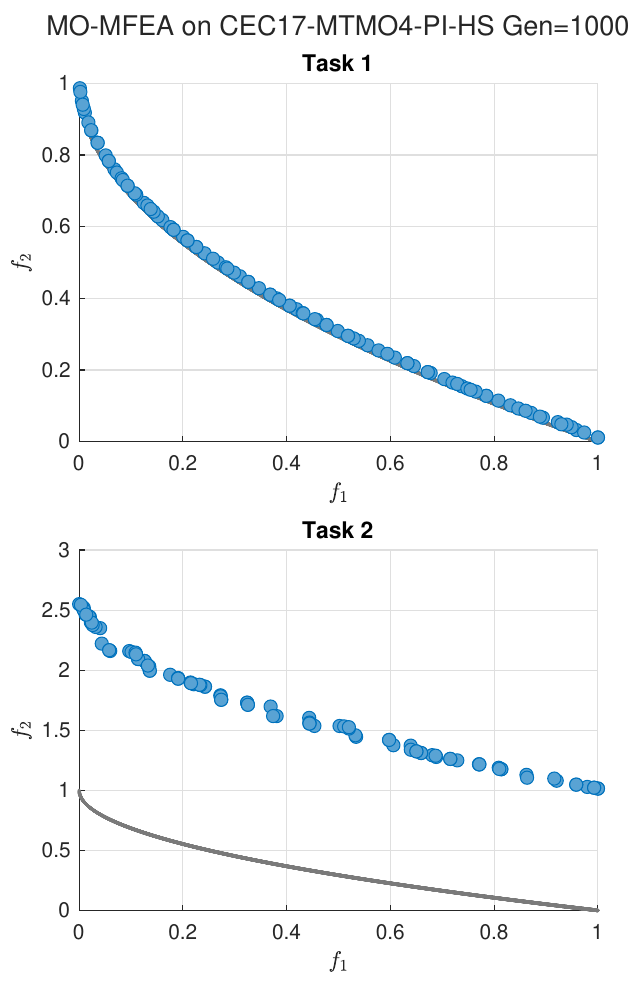}}
  \subfigure[MTDE-MKTA]{\includegraphics[scale=0.2]{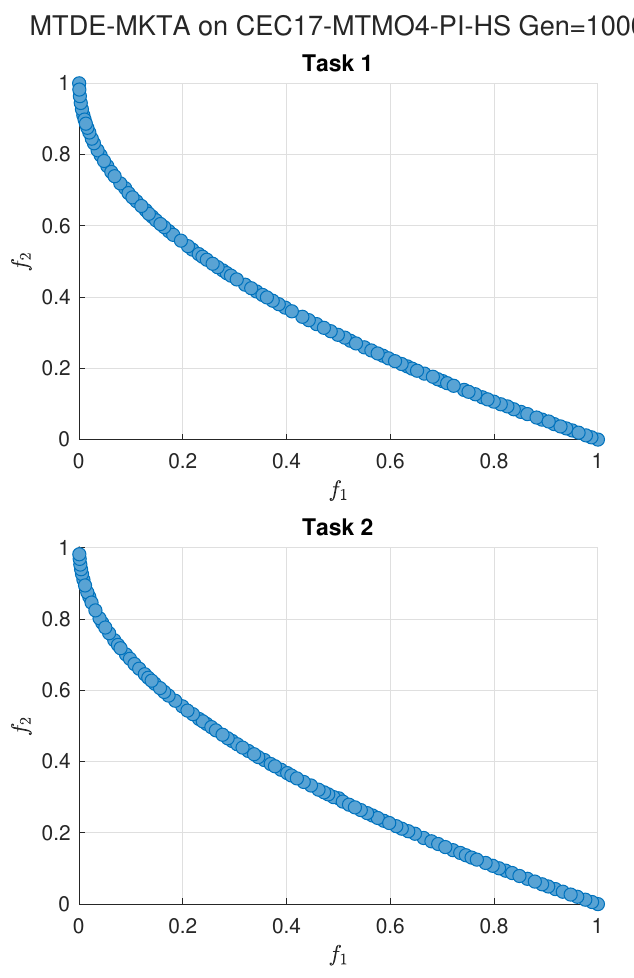}}
  \\
  \subfigure[Gen=1]{\includegraphics[scale=0.2]{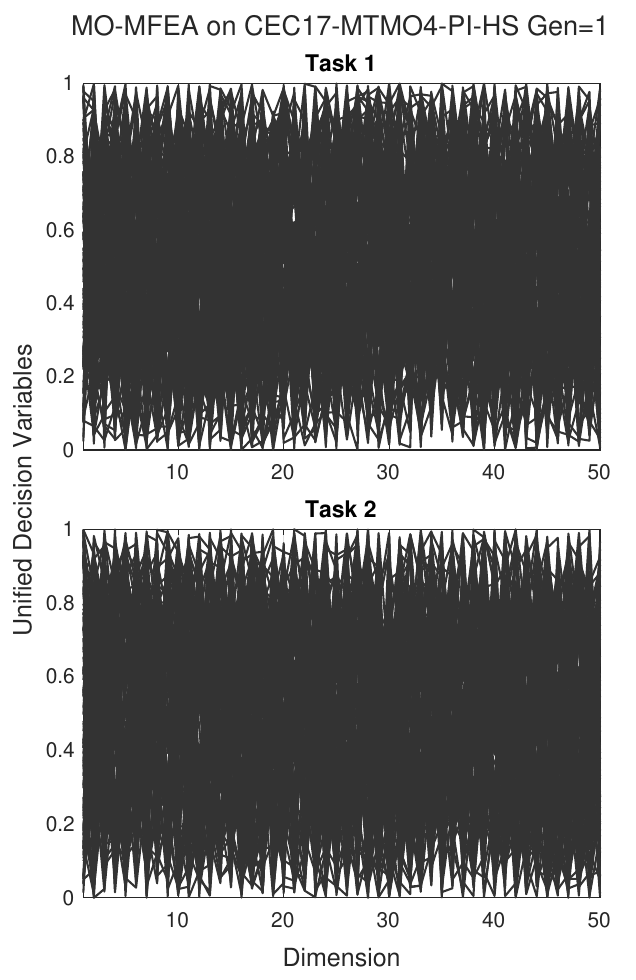}}
  \subfigure[Gen=200]{\includegraphics[scale=0.2]{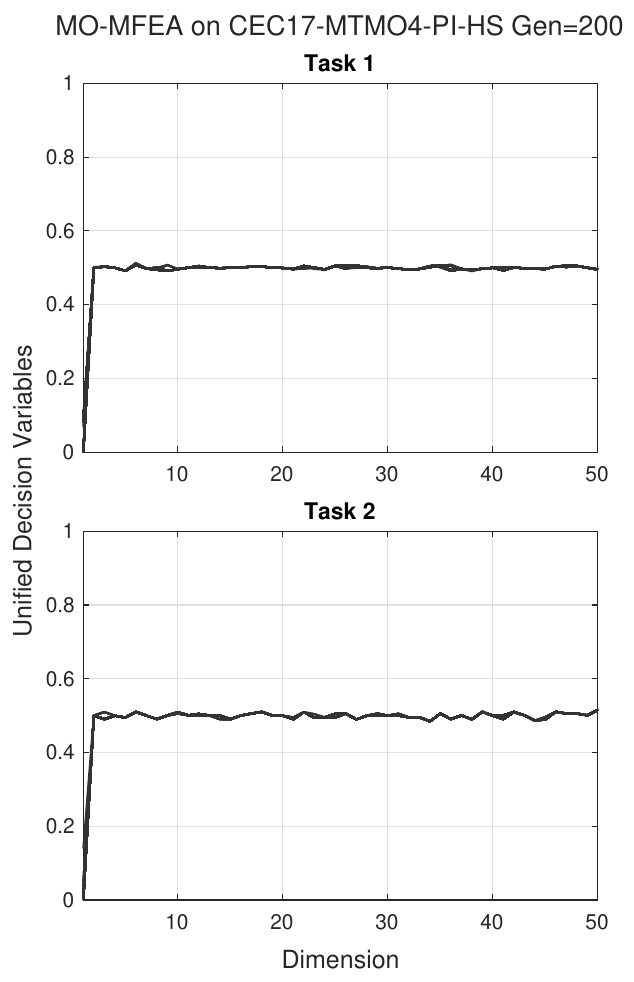}}
  \subfigure[Gen=500]{\includegraphics[scale=0.2]{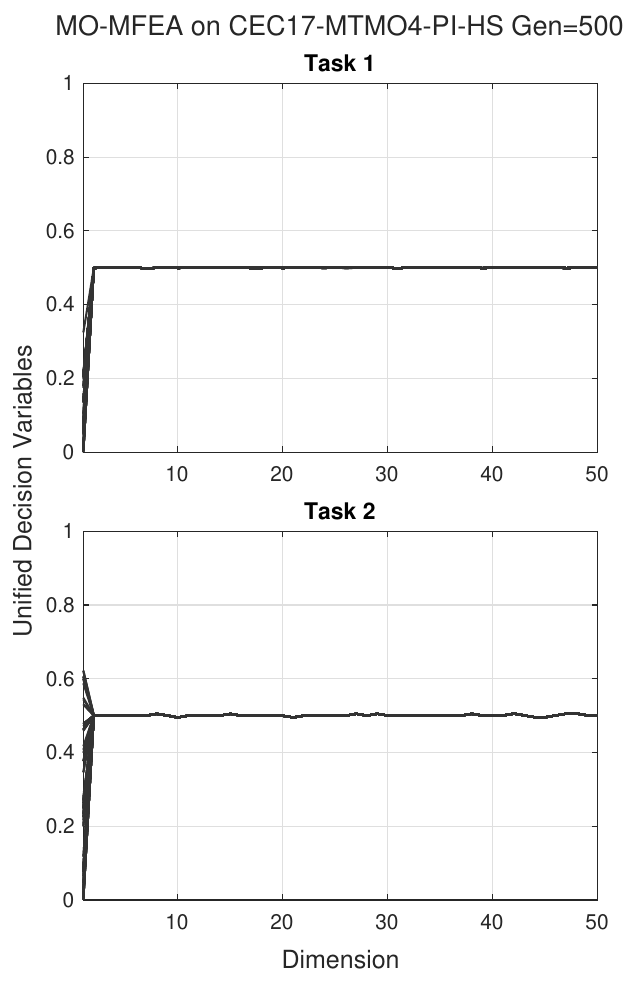}}
  \subfigure[Gen=700]{\includegraphics[scale=0.2]{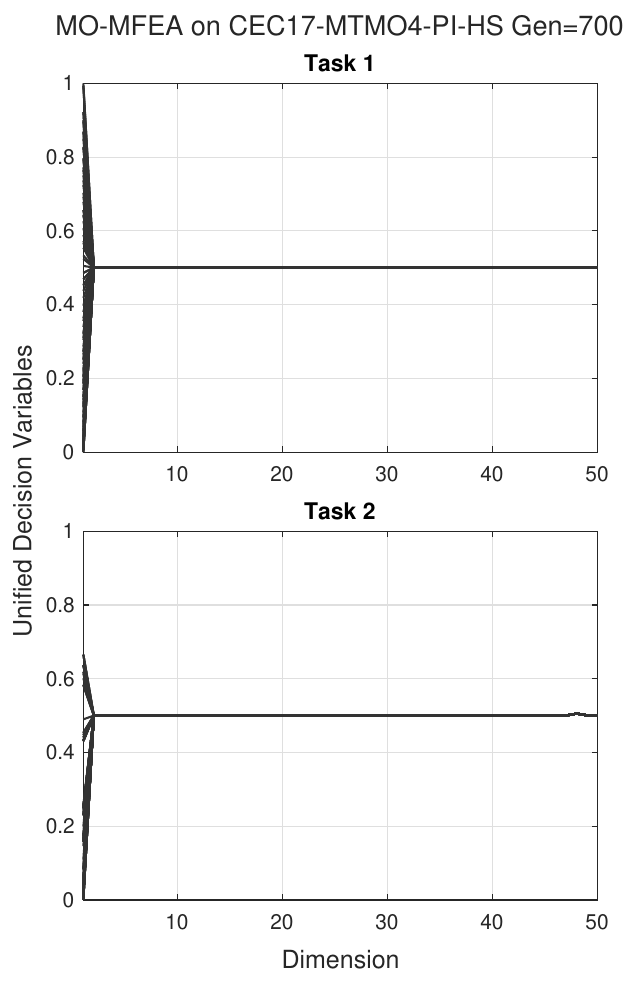}}
  \subfigure[Gen=1000]{\includegraphics[scale=0.2]{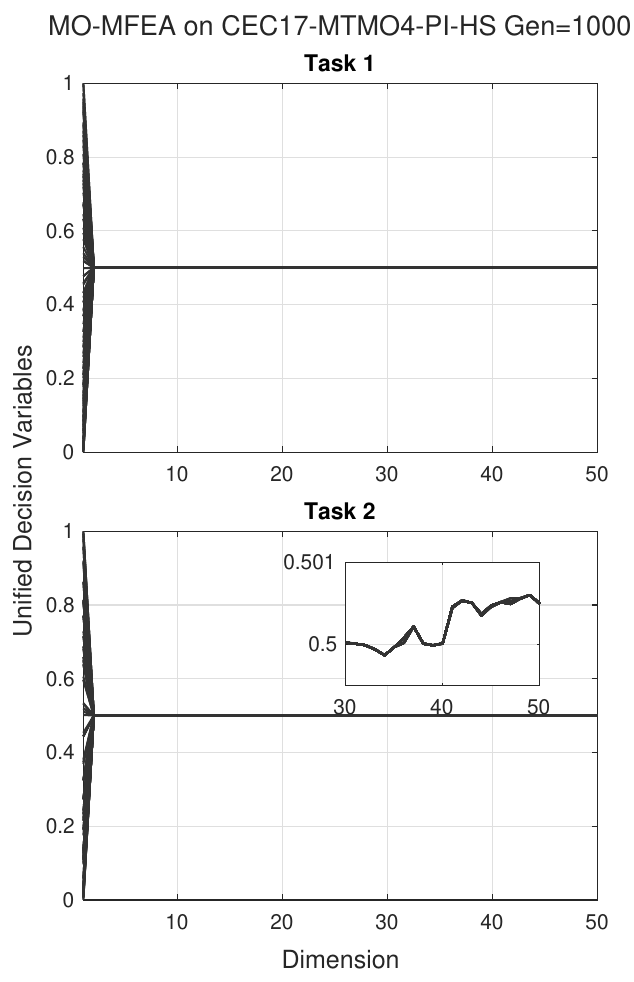}}
  \subfigure[MTDE-MKTA]{\includegraphics[scale=0.2]{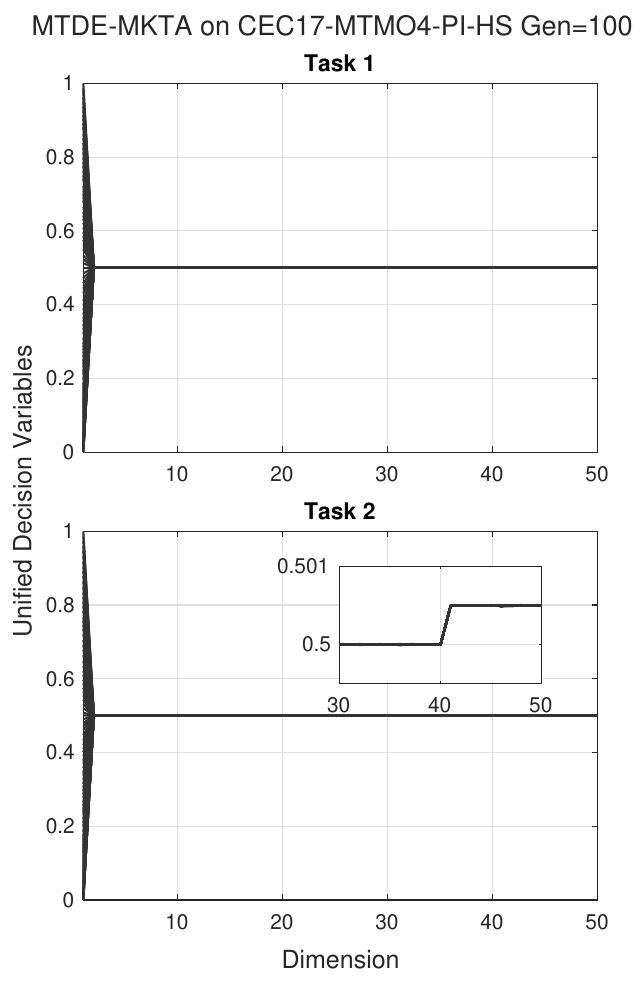}}
  \centering
  \caption{Schematic of population snapshots on CEC17-MTMO4 across generations. The top row (a)-(f) shows the objective space, where the blue dots are the algorithm's non-dominated solutions and the black line is the true {Pareto Front}. The bottom row (g)-(l) shows the 50-dimensional decision space. {The horizontal axis represents the dimension index (1 to 50), and the vertical axis represents the unified decision variable value. Each individual in the population is visualized as a polyline connecting its values across all dimensions. The dense overlap of these polylines (appearing as black bands or lines) visually reflects the convergence consistency and the value distribution of the entire population.} Panels (a)-(e) and (g)-(k) depict MO-MFEA at Gen=1, 200, 500, 700, and 1000. Panels (f) and (l) show the example of MTDE-MKTA at Gen=1000.}
  \label{fig:variation-Pareto-Dec}
\end{figure*}

{After configuring the parameters and starting the run, MToP offers dynamic display utilities such as \code{Draw Obj} and \code{Draw Dec} to enable researchers to explore algorithm behavior in real-time.} Fig.~\ref{fig:variation-Pareto-Dec} illustrates an example of population variation of MO-MFEA~\cite{Gupta2017MO-MFEA} on CEC17-MTMO4, and compares its final state with that of MTDE-MKTA~\cite{Li2025MTDE-MKTA}, showcasing the evolution in both objective space and decision space. CEC17-MTMO4 is a multi-objective problem with two tasks, each comprising two objectives. The optimal Pareto set of the first task contains diversity dimension $1$ with optimal value range $[0,1]$ and convergence dimension $[2,50]$ with optimal value $0.5$. The optimal Pareto set of the second task contains diversity dimension $1$ with optimal value range $[0,1]$, convergence dimension $[2,40]$ with optimal value $0.5$, and convergence dimension $[41,50]$ with optimal value {$0.5005$.}

  {
    Before analyzing the visual patterns, it is essential to formally distinguish the knowledge transfer mechanisms of the compared algorithms to understand the underlying causes of their behaviors.
    MO-MFEA relies on \textit{implicit knowledge transfer} via assortative mating. Let $\bm{x}_i$ and $\bm{x}_j$ be parents from different tasks. The offspring $\bm{u}_i$ is generated using the simulated binary crossover, which directly mixes the decision variables:
    \begin{equation}
      \label{eq:sbx}
      \bm{u}_{i, d} =
      \begin{cases}
        0.5[(1+\gamma)\bm{x}_{i, d} + (1-\gamma)\bm{x}_{j, d}] & \text{if } r \le 0.5 \\
        0.5[(1-\gamma)\bm{x}_{i, d} + (1+\gamma)\bm{x}_{j, d}] & \text{otherwise}
      \end{cases}
    \end{equation}
    where $\gamma$ is the spread factor and $r$ is a random number. This mechanism lacks explicit domain adaptation, which may lead to negative transfer if the optimal regions of the tasks are misaligned.

    In contrast, MTDE-MKTA employs an \textit{explicit knowledge transfer} strategy based on evolution path maintenance. It models the population distributions of the source and target tasks as Gaussian distributions $\mathcal{N}(\bm{\mu}_s, \bm{\sigma}_s)$ and $\mathcal{N}(\bm{\mu}_t, \bm{\sigma}_t)$, respectively. A solution $\bm{x}$ from the source task is explicitly transformed into a candidate solution $\bm{y}$ for the target task by aligning their statistical distributions:
    \begin{equation}
      \label{eq:evo-path-trans}
      \bm{y} = \bm{\mu}_t + \frac{\bm{\sigma}_t \circ (\bm{x} - \bm{\mu}_s)}{\bm{\sigma}_s}
    \end{equation}
    where $\circ$ denotes element-wise multiplication and division. This transformation adapts the transferred solution to the target task's search space, thereby correcting the decision variable bias and mitigating negative transfer.
  }

{Through these dynamic visualizations, researchers can intuitively grasp the algorithm's operational behavior.} {In this example, during the initial phase of evolution (Gen=1), the population initialization is dispersed across the decision space and poorly situated in the objective space. As evolution progresses through the first and middle phases {(Gen=200)}, individual gene knowledge transfer leads to populations swiftly converging to more favorable positions in the objective space, albeit at the expense of reduced diversity. Subsequently, in the middle and late stages {(Gen=500 to 700)}, the population gradually emphasizes diversity. By Gen=1000, MO-MFEA achieves the optimal {Pareto front} on the first task but becomes trapped in local optima on the second task.} {The schematics provide the key insight: the population for the second task is clustered in the similar decision space region as the population for the first task. However, the true optimal regions of these two tasks do not precisely overlap in the decision space. This observation strongly suggests the presence of negative knowledge transfer: the strong convergence of Task 1 has incorrectly pulled the Task 2 population into its own optimal region.} {This confirms the limitation of the implicit transfer mechanism (Eq.~\eqref{eq:sbx}) discussed above: without domain adaptation, the direct mixing of variables drags the population towards the wrong attractor.}

Furthermore, MToP can be used to validate the effectiveness of remedial approaches. For instance, Fig.~\ref{fig:variation-Pareto-Dec}~(f) and (l) show the final population snapshot of MTDE-MKTA, an algorithm with multiple knowledge transfer mechanisms designed to mitigate this issue. In contrast to MO-MFEA, MTDE-MKTA successfully converges to the true optimal {Pareto front} for both tasks. {This validates the efficacy of the explicit transformation (Eq.~\eqref{eq:evo-path-trans}), which successfully re-mapped the guiding solutions to the correct decision space region ($0.5005$), thus overcoming the negative transfer problem.}

\subsection{Performing Comparative Experiments}\label{sec:guide:exp}

\begin{figure*}[htbp]
  \centering
  \includegraphics[scale=0.51]{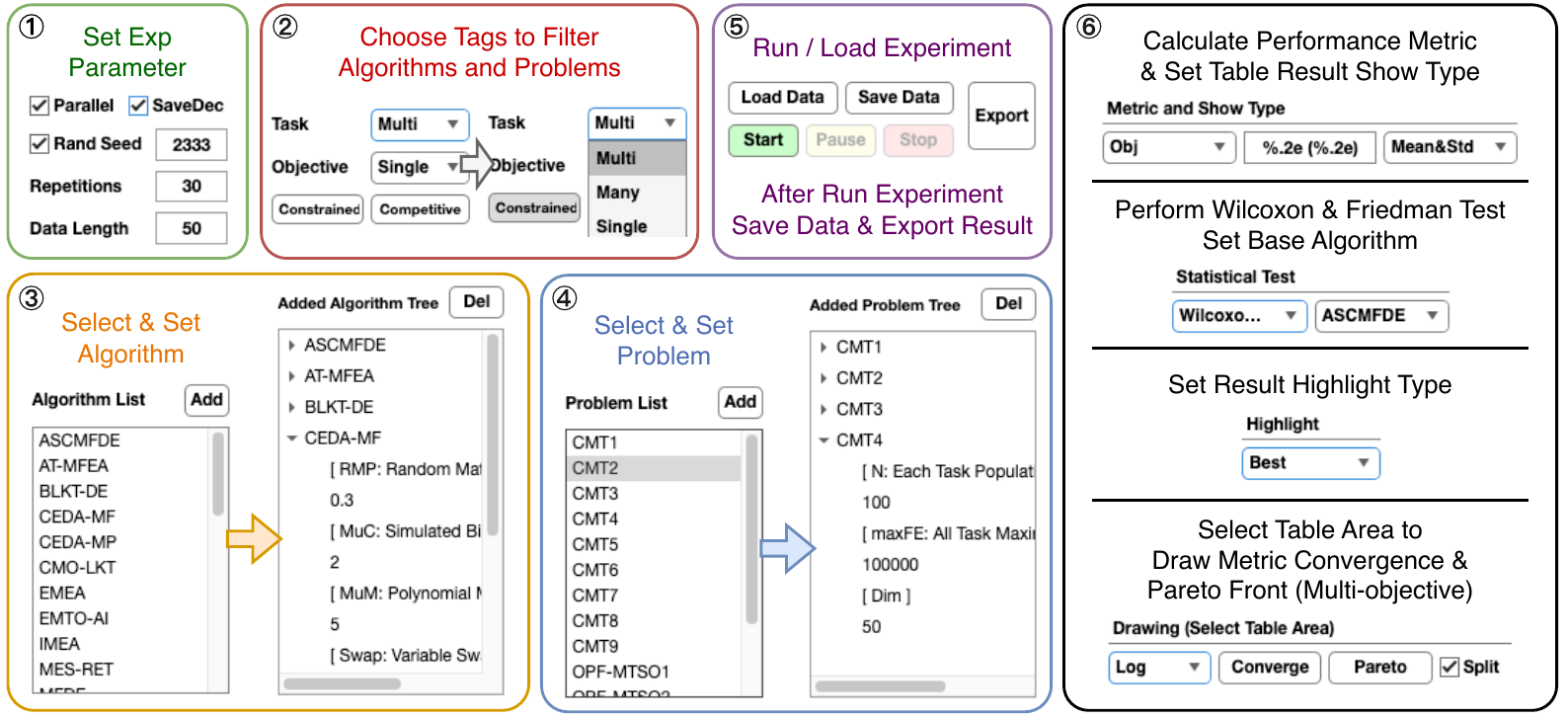}
  \centering
  \caption{Experiment execution steps in the \code{Experiment Module} of MToP GUI.}
  \label{fig:exp-flow}
\end{figure*}

The GUI of the \code{Experiment Module}, shown in Fig.~\ref{fig:mtop-module}~(b), features a three-column layout: the left and middle columns are for experimental setup, and the right column is for results analysis. {The detailed steps for conducting an experiment are illustrated in Fig.~\ref{fig:exp-flow}.}

\subsubsection{Setting Experiment Parameters}\label{sec:guide:exp:param}

{
  Experiment parameters are set in the \code{Set Exp Parameter} panel (Step 1). This includes setting the number of \code{Repetitions} (independent runs), the \code{Rand Seed} for reproducibility, and the \code{Data Length} (the number of {checkpoints} to save per run). Toggles for \code{Parallel} execution and \code{SaveDec} (to save decision variables) are also located here. Enabling \code{Parallel} allows independent runs to execute simultaneously. The number of concurrent processes depends on the system's processor cores, and MATLAB automatically handles the scheduling. The typical execution flow processes one problem at a time, running all added algorithms on it in parallel before moving to the next problem.
}

\subsubsection{Selecting Algorithms and Problems}\label{sec:guide:exp:select}

{
  As shown in Fig.~\ref{fig:exp-flow} (Steps 2-4), the user first selects the algorithms and problems to be included in the experiment. To simplify this process, users can use the tag filters (Step 2) such as \code{Task}, \code{Objective}, and \code{Constrained} to narrow down the lists based on predefined labels. From the \code{Algorithm List} (Step 3), the user selects one or more algorithms and clicks the \code{Add} button to move them to the \code{Added Algorithm Tree}. An identical process is used to select problems from the \code{Problem List} and add them to the \code{Added Problem Tree} (Step 4).

  As shown in Steps 3 and 4, users can set parameters for specific algorithms and problems. By clicking on an item in the \code{Added Algorithm Tree} or \code{Added Problem Tree}, its individual parameters (\eg, population size \code{N}, operator parameters \code{RMP}) are displayed in the text box below, where they can be directly edited.

  To assist researchers in this selection, especially when starting a new study, Table~\ref{tab:recommendations} provides a curated guide to benchmark suites and high-performing algorithms. The listed benchmarks are widely used and recognized within the EMT community, serving as standard testbeds. The recommended algorithms are not arbitrary; they represent the top 5 algorithms based on the Friedman ranking of their average metric performance over 30 independent runs on each benchmark set. The full experimental validation for this ranking is detailed later in Section~\ref{sec:validate:repro}. This table serves as an excellent starting point for researchers and can also offer valuable insights when selecting algorithms for new real-world problems. Comprehensive lists, references, and detailed descriptions of all problems, algorithms, and metrics available in MToP are provided in Table~\ref{tab:prob-mt}, Table~\ref{tab:algo-mt}, {and Table~\ref{tab:metric}, respectively.}
}

\begin{table*}[htbp]
  \centering
  \renewcommand{\arraystretch}{1.1}
  \caption{Recommended benchmarks, algorithms and metrics for different MTO problem categories in MToP. The selected algorithms are based on their Friedman rankings on each benchmark. Detailed references and descriptions of these algorithms and benchmarks can be found in supplementary files.}
  \label{tab:recommendations}
  \resizebox{\textwidth}{!}{
    \setlength{\tabcolsep}{3mm}{
      \begin{tabular}{lll}
        \toprule
        Benchmark    & Recommended Algorithms (selected by Friedman rankings)    & Recommended Metric(s) \\
        \midrule
        \multicolumn{3}{l}{\textit{\textbf{Single-objective Multi-task}}}                                \\
        CEC17-MTSO   & MFEA-GHS, MTES-KG, MFMP, MTDE-ADKT, MTEA-HKTS             & Obj                   \\
        WCCI20-MTSO  & MTES-KG, MFMP, MTDE-ADKT, MTEA-PAE, MFEA-DGD              & Obj                   \\
        \midrule
        \multicolumn{3}{l}{\textit{\textbf{Constrained Single-objective Multi-task}}}                    \\
        CMT          & MTEA-PAE, MTES-KG, CEDA-MP, MFEA-GHS, MTEA-AD             & Obj, CV               \\
        \midrule
        \multicolumn{3}{l}{\textit{\textbf{Competitive Single-objective Multi-task}}}                    \\
        C2TOP\&C4TOP & MTSRA, MFMP, MTES-KG, MTEA-HKTS, DEORA                    & Obj (CMT)             \\
        C-CPLX       & MTES-KG, MFMP, MTSRA, MTEA-PAE, MTEA-HKTS                 & Obj (CMT)             \\
        \midrule
        \multicolumn{3}{l}{\textit{\textbf{Single-objective Many-task}}}                                 \\
        CEC19-MaTSO  & DTSKT, TNG-SNES, MTES-KG, MTEA-HKTS, SBCMAES              & Obj (MTS), Obj (UV)   \\
        WCCI20-MaTSO & DTSKT, TNG-SNES, MTES-KG, SBCMAES, EMaTO-MKT              & Obj (MTS), Obj (UV)   \\
        \midrule
        \multicolumn{3}{l}{\textit{\textbf{Multi-objective Multi-task}}}                                 \\
        CEC17-MTMO   & MO-MTEA-PAE, MTDE-MKTA, MTEA-DCK, MTEA-D-DN, KR-MTEA      & IGD+, HV              \\
        CEC19-MTMO   & MTDE-MKTA, MTEA-DCK, MTEA-D-TSD, MO-MTEA-PAE, MTEA-D-DN   & IGD+, HV              \\
        CEC21-MTMO   & MTEA-D-TSD, MO-MTEA-PAE, MTEA-DCK, MO-MTEA-SaO, MTDE-MKTA & IGD+, HV              \\
        \midrule
        \multicolumn{3}{l}{\textit{\textbf{Competitive Multi-objective Multi-task}}}                     \\
        CMOMT        & RVC-MTEA, MO-MTEA-PAE, MO-MTEA-SaO,  MO-EMEA, MO-MFEA-II  & IGD+ (CMT), HV (CMT)  \\
        \bottomrule
      \end{tabular}}}
\end{table*}

\subsubsection{Running Experiment and Calculating Metrics}\label{sec:guide:exp:run}

{
  Once all configurations are set, the experiment is initiated by clicking the \code{Start} button in the \code{Run / Load Experiment} panel (Step 5). The \code{Pause} and \code{Stop} buttons can be used to control the execution. After the experiment completes, the user can move to the results panel (Step 6) to analyze the performance. The first step is to select a \code{Metric} (\eg, \code{Obj}) and a \code{Show Type} (\eg, \code{Mean\&Std}). MToP then calculates the results and populates the main table.

  Table~\ref{tab:recommendations} also provides guidance on which metrics are most appropriate, as the choice is critical for correct analysis:
  \begin{itemize}
    \item For standard \textit{multi-task} problems (\eg, CEC17-MTSO, CEC19-MTMO), the number of tasks is small, making it practical and recommended to analyze per-task metrics directly, such as \code{Obj}, \code{IGD+}, or \code{HV}. For multi-objective metrics, {\code{IGD+} is generally recommended over the traditional \code{IGD} because, unlike \code{IGD}, it is weakly Pareto-compliant and therefore provides a more comprehensive and stable measure of both convergence and distribution.} Furthermore, the \code{HV} metric is also provided, as it is a valuable indicator that does not require a known true Pareto front, making it suitable for real-world problems where the optimum is unknown.
    \item {For \textit{constrained multi-task} problems (\eg, CMT), the evaluation must account for both feasibility and objective quality. The standard approach is to report the \code{Obj} (objective value) for runs that find at least one feasible solution, but to report the \code{CV} (constraint violation) for runs that fail to find any feasible solutions.}
    \item For \textit{competitive multi-task} problems (\eg, C-CPLX, CMOMT), per-task analysis is not correct, and aggregated metrics like \code{Obj (CMT)} or \code{IGD+ (CMT)} are necessary to capture the overall trade-off across tasks.
    \item For \textit{many-task} problems (\eg, CEC19-MaTSO), the large number of tasks makes reporting per-task metrics impractical. Here, we recommend using widely-adopted summary metrics such as the multi-task score \code{Obj (MTS)} (relative performance across tasks) or the unified value \code{Obj (UV)} (adjusted absolute performance across tasks).
    \item For all problem types, the \code{Run Time} metric can be used to compare the computational efficiency and execution time of different algorithms.
  \end{itemize}
}

\subsubsection{Performing Statistical Tests}\label{sec:guide:exp:stat}

{
  The results panel (Step 6) also includes a suite for statistical analysis. MToP provides two distinct non-parametric approaches for performance comparison: direct pairwise tests and a global omnibus test with post-hoc analysis.

  First, for direct pairwise comparisons, MToP offers the \textbf{Wilcoxon rank-sum test} and the \textbf{Wilcoxon signed-rank test}. When a user selects one of these methods, MToP performs a direct statistical comparison between each algorithm and a user-selected \code{Base Algorithm}. MToP's logic then annotates the results table with ``+'' (significantly better), ``-'' (significantly worse), or ``='' (no significant difference) based on a $p$-value threshold of $0.05$. It is important to note that these are uncorrected pairwise tests intended for exploratory analysis.

  Second, for a more robust global comparison, MToP provides the \textbf{Friedman test}. This is an omnibus test that first checks for any statistically significant differences among the entire set of selected algorithms. Users can choose to run this test on the mean results across all runs \code{(mean)} or on the complete dataset combining all runs \code{(all reps)}. Upon selection, MToP automatically performs the global Friedman test and then immediately conducts a post-hoc test. This post-hoc analysis calculates the $z$-value and corresponding $p$-value to compare the mean rank of each algorithm against the mean rank of the \code{Base Algorithm}. The results table is then populated with the mean rank for each algorithm and the $p$-value for its comparison against the base, allowing for a statistically grounded, rank-based assessment.

  It is a known characteristic of null-hypothesis significance testing that with a sufficiently large number of replications, even trivial performance differences can be deemed statistically significant. This does not necessarily imply practical significance. MToP is therefore designed to facilitate a multi-faceted analysis. MToP provides both relative performance comparisons (via the Wilcoxon and Friedman statistical test results) and crucial absolute performance measures (the \code{Mean\&Std} values). Researchers are thus equipped to interpret the statistical results in conjunction with the absolute performance data and the MToP's visualizations (\eg, \code{Converge} plots) to make a more informed judgment about the practical significance of the observed differences.
}

\subsubsection{Exporting Experiment Data and Results}\label{sec:guide:exp:export}

{

  MToP provides two primary methods for saving data, located in the \code{Run / Load Experiment} panel (Fig.~\ref{fig:exp-flow}, Step 5). The \code{Save Data} and \code{Load Data} buttons are used to save or reload the entire experimental session (the raw \code{MTOData} object), which allows for experiments to be reloaded and analyzed later.

  For saving processed results, the \code{Export} button provides more granular control. Clicking this button opens a new dialog window offering three distinct export options:
  \begin{itemize}
    \item \textbf{\code{Current Table (tex, xlsx, csv)}:} This option saves the data exactly as it is currently displayed in the main results table, including the calculated metrics and any statistical annotations (\eg, ``+ / - / ='' and ``Ranking''). It supports \code{tex}, \code{xlsx}, and \code{csv} formats for easy integration into publications and spreadsheets.
    \item \textbf{\code{IOHanalyzer Data (csv)}:} This option exports the complete convergence history (not just the final values) for all selected runs into a \code{csv} format. This format is specifically structured for use with the IOHanalyzer~\cite{Wang2022IOHanalyzer}, facilitating advanced external analysis.
    \item \textbf{\code{Best Decision Variable (mat)}:} This option extracts the decision variables of the best-found solutions of all algorithms and repetitions and saves them to a \code{.mat} file for further inspection or use in other applications. For single-objective MTO problems, the best solution is determined by the lowest objective value. For multi-objective MTO problems, the best solutions are a set of Pareto sets~\cite{Liu2024Finding,Choong2022Jack}.
  \end{itemize}
}

\subsubsection{Visualizing Metric Convergence and Pareto Front}\label{sec:guide:exp:visual}

{
  Finally, the module offers integrated plotting tools at the bottom of the results panel (Step 6). By selecting one or more rows and columns in the results table, the user can generate plots. Clicking the \code{Converge} button generates a convergence plot for the selected metric results, with a \code{Log} toggle available to switch the y-axis to a logarithmic scale, and a \code{Range} toggle to display the 95\% confidence interval around the mean curve. For multi-objective problems, clicking the \code{Pareto} button will plot the final acquired non-dominated solutions ({Pareto front approximations}). Additionally, the convergence data can be exported in a format compatible with IOHanalyzer~\cite{Wang2022IOHanalyzer} for further external analysis and visualization.
}

\subsection{Processing Experiment Data}\label{sec:guide:data}

{ MToP provides a dedicated \code{Data Process Module} within the GUI for merging and splitting experimental data files. This module allows users to combine multiple \code{MTOData.mat} files into a single comprehensive dataset or to partition a large dataset into smaller, more manageable segments based on specific criteria such as algorithms, problems, or repetitions. The merging function is particularly useful for aggregating results from separate experimental runs, while the splitting function aids in isolating subsets of data for focused analysis. Users can access this module directly from the main GUI, where they can select the desired operation (merge or split), specify input files, and define output parameters. The processed data is saved in the standard \code{MTOData.mat} format, ensuring compatibility with other modules within MToP. }

\begin{figure*}[htbp]
  \centering
  \includegraphics[scale=0.45]{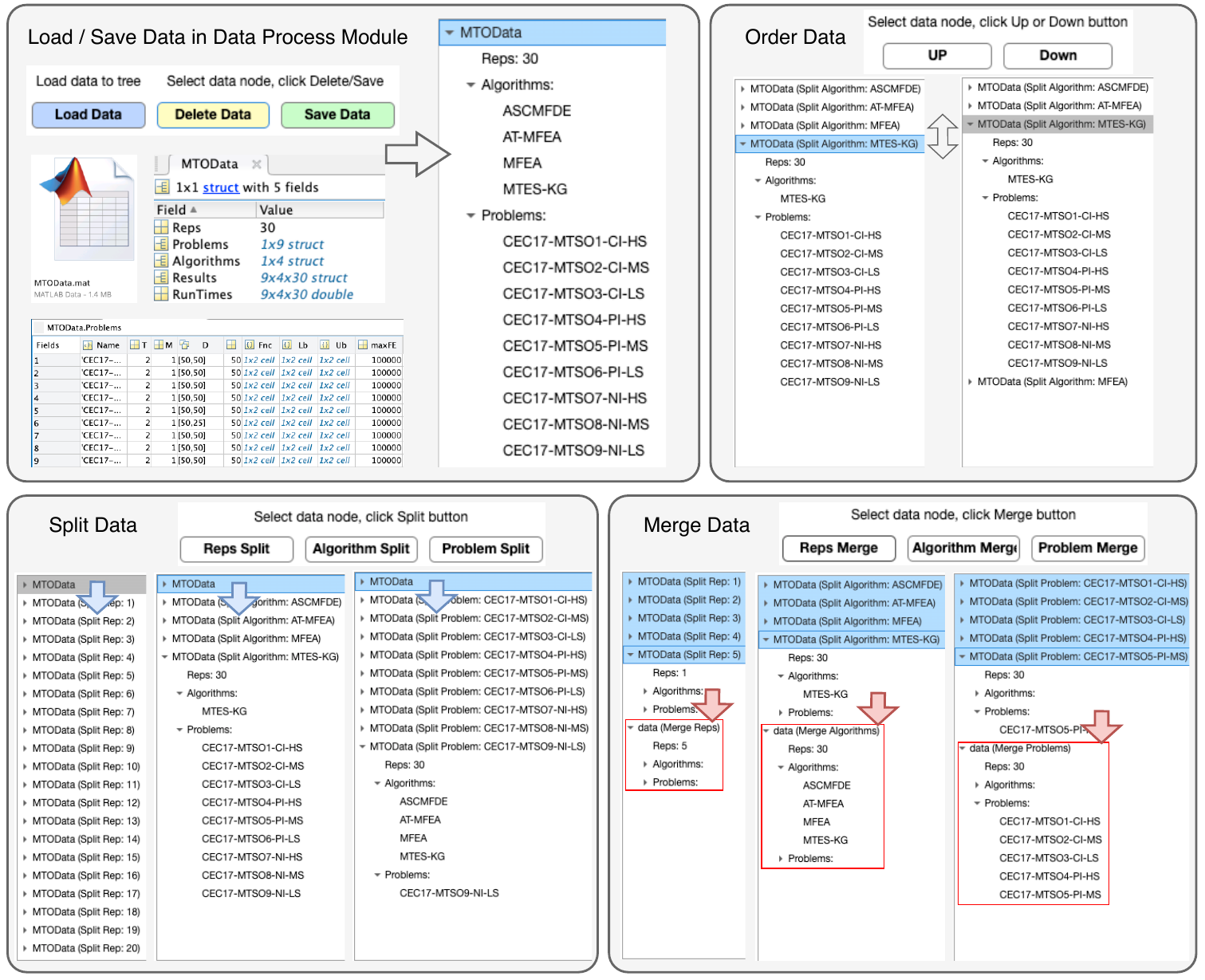}
  \centering
  \caption{Data processing steps in the \code{Data Process Module} of MToP GUI.}
  \label{fig:data-process}
\end{figure*}

\subsubsection{Loading and Viewing Data Files}\label{sec:guide:data:load}

{
  As shown in the \code{Load / Save Data} panel of Fig.~\ref{fig:data-process}, the user begins by clicking the \code{Load Data} button. This opens a file dialog to select one or more \code{MTOData.mat} files, which are then added to the \code{Data Tree}. By clicking on a loaded data object, its high-level structure (\eg, \code{Reps}, \code{Algorithms}, \code{Problems}) is displayed. Users can further click on these fields, such as \code{Problems}, to view the detailed contents in a table, allowing for verification of the data before processing. The \code{Delete Data} button can also be used to remove any unwanted or intermediate data objects from the \code{Data Tree}.
}

\subsubsection{Splitting Experiment Data}\label{sec:guide:data:split}

{
  The \code{Split Data} panel in Fig.~\ref{fig:data-process} illustrates the partitioning functionality. To divide a large dataset, the user selects a data object from the \code{Data Tree}. MToP provides three splitting options: \code{Reps Split}, \code{Algorithm Split}, and \code{Problem Split}. For instance, selecting \code{Algorithm Split} will parse the dataset and create new, separate data objects for each algorithm it contains (\eg, \code{MTOData (Split Algorithm: MFEA)}). These new objects then appear in the \code{Data Tree}, ready for isolated analysis or saving.
}

\subsubsection{Merging Experiment Data}\label{sec:guide:data:merge}

{
  Conversely, the \code{Merge Data} panel demonstrates how to combine datasets. The user selects two or more data objects from the \code{Data Tree} that share compatible settings. For example, to aggregate independent runs, the user can select multiple files and click \code{Reps Merge}. This creates a single new dataset (\eg, \code{Data (Merge Reps)}) with a larger total number of repetitions. The \code{Algorithm Merge} and \code{Problem Merge} buttons function similarly, allowing for the consolidation of data across algorithms or problems. Note that when merging, MToP checks for consistency in experiment settings to ensure valid combinations (\eg, same problems and reps when merging algorithms).
}

\subsubsection{Adjusting Data Precision}\label{sec:guide:data:precision}

{
  The module also provides tools for changing data precision. Users can select a data object and specify the desired number of decimal places for metric values. This is particularly useful for reducing file size or standardizing data formats before saving.
}

\subsubsection{Saving Processed Data}\label{sec:guide:data:save}

{
  After all processing operations (splitting, merging, or reordering) are complete, the new or modified data objects can be exported. As shown in the \code{Load / Save Data} panel, the user selects the desired data object from the \code{Data Tree} (\eg, \code{Data (Merge Reps)}) and clicks the \code{Save Data} button to save it as a new \code{MTOData.mat} file.
}

\subsection{Executing via Alternative Command Line}\label{sec:guide:cmd}

{
  For users who require batch processing, integration with external scripts, or operation in non-GUI environments, MToP provides a comprehensive and flexible command-line interface. The main \code{mto.m} script has been created to serve as a dual-purpose entry point: calling \code{mto} with no arguments launches the GUI, while calling it with arguments executes a command-line experiment.

  The function \code{MTOData = mto(varargin)} is designed for maximum flexibility, accepting both positional arguments and Name-Value pairs. For instance, users can run basic experiments using commands like \code{result = mto(MFEA, CMT1)} for quick runs. More complex configurations such as \code{mto(\{MFEA, EMEA\}, \{CMT1, CMT2\}, 'Reps', 30, 'Par\_Flag', true, 'Save\_Name', 'A.mat')} allow for detailed customization, including parallel execution and specifying output filenames. Upon completion, the function returns the standard \code{MTOData} struct and simultaneously saves it to the specified \code{.mat} file. This command-line interface workflow extends beyond just running experiments; the returned \code{MTOData} object can be directly passed to any metric function to calculate results, just as in the GUI. This enables a complete, script-based workflow: setup, run, and analyze.

  A comprehensive set of examples demonstrating this full process including how to instantiate and modify algorithm/problem objects, run experiments, calculate metrics like \code{Obj} and \code{IGD+}, and programmatically plot the final convergence curves and {Pareto front approximations} is provided in the \code{cmd\_examples.m} script located in MToP's root directory.
}

\section{Validation and Comparison}\label{sec:validate}

{
  This section first performs reproducible validation tests to ensure that the algorithms and problems implemented in MToP are reliable and accurate. Following this, a performance comparison between MToP and a popular EC platform, PlatEMO~\cite{Tian2017PlatEMO}, is conducted to demonstrate MToP's comparable efficiency.
}

\subsection{Reproducible Validation}\label{sec:validate:repro}

{
  A core requirement for a scientific platform is the reliability and reproducibility of its components. To validate the implementations of the extensive algorithm and problem libraries within MToP, we conducted a comprehensive experimental study, running all algorithms (including both MTEAs and single-task EAs) on all compatible benchmark suites for 30 independent runs.

  To ensure fair comparisons and high reproducibility, a global random seed controller is employed. For each independent repetition \code{rep}, the random number generator is explicitly seeded using \code{rng(Global\_Seed + rep - 1)}. This ensures that all algorithms within the same repetition begin with identical random conditions, which is crucial for reliable comparisons. In order to enhance reproducibility for the wider community and to provide a baseline for future research, we have made the complete raw experimental data from this large-scale validation publicly available. This dataset validates our implementations and can be used directly by other researchers to avoid redundant, computationally expensive experiments. All data can be accessed and downloaded.\footnote{Pre-run experiment data can be found at:\\
    \hspace*{1em} Google Drive: \href{https://drive.google.com/drive/folders/1IpwXNuOYcnpC3IXbAx3VnLGVV899k3bG?usp=share_link}{\color{violet}\tiny\texttt{https://drive.google.com/drive/folders/1IpwXNuOYcnpC3IXbAx3VnLGVV899k3bG?usp=share\_link}} \\
    \hspace*{1em} Hugging Face: \href{https://huggingface.co/datasets/intLyc/MToP-MTOData/tree/main}{\color{violet}\tiny\texttt{https://huggingface.co/datasets/intLyc/MToP-MTOData/tree/main}} \\
    \hspace*{1em} Baidu Netdisk: \href{https://pan.baidu.com/s/1Pk06fBj_4gidkiZe4f1Oww?pwd=mtop}{\color{violet}\tiny\texttt{https://pan.baidu.com/s/1Pk06fBj\_4gidkiZe4f1Oww?pwd=mtop}}
  }
}

\subsection{Comparison with Other Platforms}\label{sec:validate:compare}

\begin{figure}[htbp]
  \centering
  \includegraphics[scale=0.55]{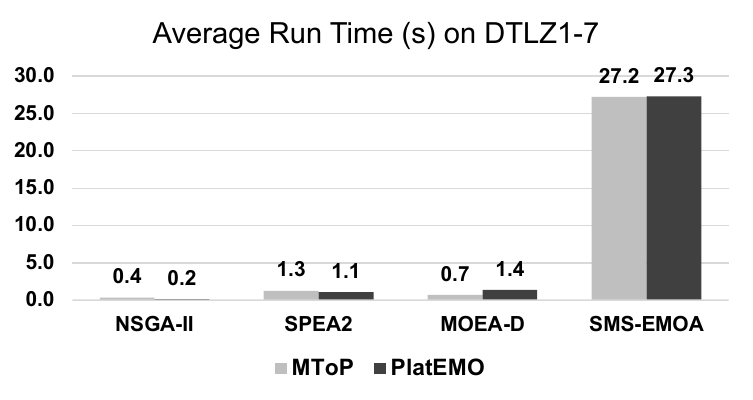}
  \centering
  \caption{Running time comparison between MToP and PlatEMO on DTLZ problems using algorithms: NSGA-II, SPEA2, MOEA-D, and SMS-EMOA.}
  \label{fig:comparison}
\end{figure}

{
Beyond internal validation, it is crucial to benchmark MToP's performance and efficiency against established, state-of-the-art platforms. We selected PlatEMO~\cite{Tian2017PlatEMO}, a widely-used and highly-regarded platform for evolutionary computation, as the basis for our comparison. To ensure a fair and direct comparison, we selected four popular and well-understood single-task multi-objective algorithms that are implemented in both platforms: NSGA-II, SPEA2, MOEA-D, and SMS-EMOA. We executed these algorithms on the standard DTLZ1-7 benchmark suite. Both platforms were run under identical experimental settings, including 30 independent runs and the use of parallel execution.

First, we compared the computational efficiency. Fig.~\ref{fig:comparison} displays the average wall-clock run time for each algorithm across the entire DTLZ suite. The results show that the execution times for MToP and PlatEMO are highly comparable. For NSGA-II, SPEA2, and MOEA-D, the run times are nearly identical. For SMS-EMOA, which involves computationally intensive hypervolume calculations, both platforms show similar overhead. This demonstrates that MToP's architecture is lightweight and efficient, with no significant computational overhead compared to PlatEMO.

Second, we validated the correctness of our algorithmic implementations by comparing the final solution quality. {Table~\ref{tab:results-dtlz} presents the mean and standard deviation of the \code{IGD+} metric for 30 runs. A Wilcoxon rank-sum test (at a 0.05 significance level) was conducted between the MToP and PlatEMO results for each case. The \code{+ / - / =} summary shows that there is no statistically significant difference in the vast majority of cases (24 out of 28).} The few minor differences are likely due to small implementation-level variations, but the overall performance is statistically indistinguishable. This similarity in performance is further confirmed visually in Fig.~\ref{fig:pareto-comparison}, which plots the final {Pareto front approximations} from a sample run for each algorithm. The acquired fronts from both platforms are visually coincident.
}

\section{Discussion and Outlook}\label{sec:discuss}

This paper introduces MToP, an open-source MATLAB platform designed for EMT. MToP features a user-friendly GUI, a rich collection of algorithms and problems, and convenient code structures. The current version of MToP contains over 50 MTEAs, more than 50 single-task EAs capable of handling MTO problems, over 200 benchmark MTO problems, and several real-world applications of EMT. {This paper provides recommendations for selecting algorithms and benchmarks on different types of MTO problems. It also offers guidelines for using MToP, including testing algorithms and problems, executing comparative experiments, processing experiment data, and running via the command line. Finally, this paper validates the reproducibility of implemented algorithms and problems in MToP, provides open-sourced experimental data, and compares the running time and results of MToP with other EC platforms.}

While MToP has undergone careful reimplementation, modification, and testing of included algorithms and problems, it may still contain some implementation errors and bugs. Continuous efforts are underway to identify and rectify such issues, with updates regularly posted on GitHub. MToP encourages feedback and contributions from users and researchers, with many codes already received and integrated from the community.

Moving forward, the development and enhancement of MToP will continue, drawing inspiration and contributions from the academic community. It is our aspiration that MToP evolves into a valuable tool to propel research in the field of EMT and, by extension, advance the broader evolutionary computation field. {Specifically, MToP is considering the incorporation of many-objective optimization codes~\cite{Cheng2016RVEA, Deb2014NSGA3}, multitask combinatorial optimization codes~\cite{Feng2021VRPHTO, Feng2021EEMTA} and more real-world applications~\cite{Gupta2021Half, Huang2023Evolutionary} in the future.}

\begin{acks}
  This work was partly supported by the National Natural Science Foundation of China (Grant No. 62076225), the Major Research \& Development Program of Hubei Province (Grant No. 2025BEB002 \& 2023BCA006), the Regional Innovation System Program of Hubei Province (Grant No. 2025EIA022), and the Hubei Superior and Distinctive Discipline Group of ``New-Energy Vehicles and Smart Transportation''.

  All algorithm, problem, and metric implementations within MToP adhere closely to the methodologies outlined in their respective articles. The code provided by the original authors undergoes modifications based on the source code during its integration into MToP.

  The authors would like to thank the researchers who assisted in modifying the code into the MToP pattern and the researchers who open-sourced their codes. Some of the function implementations in MToP about multi-objective optimization are referred to the code implementation in PlatEMO~\cite{Tian2017PlatEMO}, for which the authors are also thankful.
\end{acks}

\bibliographystyle{ACM-Reference-Format}
\bibliography{Reference}

\end{document}